\pdfoutput=1

\documentclass[10pt, journal,twoside]{IEEEtran}

\usepackage{xspace}

\makeatletter
\def\onedot{\ifx\@let@token.\else.\null\fi\xspace}
\def\ie{\emph{i.e}\onedot}
\makeatother

\ifCLASSINFOpdf
\else
\fi

\usepackage{cite}
\usepackage{amsmath}
\usepackage{amsfonts}
\usepackage{algorithm}
\usepackage{algorithmic}
\usepackage{threeparttable}
\usepackage{graphicx}
\usepackage{subfigure}
\usepackage{color}
\usepackage{float}
\usepackage{booktabs}
\usepackage{makecell}
\usepackage{multirow}
\usepackage{pifont}
\usepackage{amssymb}
\usepackage{mathrsfs}

\usepackage{listings}
\usepackage{xcolor}
\usepackage{graphicx}
\usepackage{amsmath}
\usepackage{amssymb}
\usepackage{booktabs}
\usepackage{multirow}
\usepackage[table]{xcolor}

\usepackage[pagebackref,breaklinks,colorlinks]{hyperref}

\usepackage{pifont}

\usepackage{framed}
\usepackage{subcaption}

\begin{document}

%
\title{GeoZero: Incentivizing Reasoning from Scratch on Geospatial Scenes}
%
%
%

\author{
Di~Wang,
Shunyu~Liu,
Wentao~Jiang,
Fengxiang~Wang,
Yi~Liu,
Xiaolei~Qin,
Zhiming~Luo,
Chaoyang~Zhou,
Haonan~Guo,
Jing~Zhang,~\IEEEmembership{Senior Member,~IEEE}
Bo~Du,~\IEEEmembership{Senior Member,~IEEE}\\
Dacheng~Tao,~\IEEEmembership{Fellow,~IEEE}
Liangpei~Zhang,~\IEEEmembership{Fellow,~IEEE}
\thanks{Corresponding authors: Jing Zhang, Bo Du, and Liangpei Zhang.}
\thanks{Di Wang, Wentao Jiang, Yi Liu, Xiaolei Qin, Zhiming Luo, Chaoyang Zhou, Haonan Guo, Jing Zhang, Bo Du, and Liangpei Zhang are with the Wuhan University, Wuhan 430072, China. Yi Liu and Zhiming Luo are also with the Zhongguancun Academy, Beijing 100094, China (e-mail: d\_wang@whu.edu.cn; jingzhang.cv@gmail.com; dubo@whu.edu.cn; zlp62@whu.edu.cn).}
\thanks{Shunyu Liu and Dacheng Tao are with the Nanyang Technological University, Singapore.}
\thanks{Fengxiang Wang is with the Shanghai AI Laboratory, Shanghai, China.}

}

%
%

\markboth{\LaTeX\ Class Files,~Vol.~X, No.~X, XXX XXX}%
{Shell \MakeLowercase{\textit{et al.}}: Bare Demo of IEEEtran.cls for IEEE}
%



\maketitle

\begin{abstract}

Multimodal large language models (MLLMs) have undergone rapid development in advancing geospatial scene understanding. Recent studies have sought to enhance the reasoning capabilities of remote sensing MLLMs, typically through cold-start training with elaborately curated chain-of-thought (CoT) data. However, this approach not only incurs substantial annotation costs but also introduces human biases that may limit the diversity of model reasoning. To address these challenges, we propose GeoZero, a framework that enables MLLMs to perform geospatial reasoning without any predefined CoT supervision. Specifically, we construct two datasets, GeoZero-Instruct and GeoZero-Hard. GeoZero-Instruct allows the model to acquire preliminary geospatial knowledge through supervised fine-tuning, while GeoZero-Hard stimulates deep reasoning during the subsequent reinforcement learning stage. Furthermore, we introduce Answer-Anchored Group Relative Policy Optimization (A$^2$GRPO), where the reasoning process is regularized by the model’s own answers, encouraging diverse yet accurate thinking. Extensive experiments on multiple remote sensing vision–language benchmarks demonstrate that GeoZero not only surpasses existing state-of-the-art methods but also fosters universal emergent reasoning capabilities across diverse geospatial tasks. Code, data, and models are available at \href{https://github.com/MiliLab/GeoZero}{GeoZero}.

\end{abstract}


%
\IEEEpeerreviewmaketitle

\section{Introduction}

Geospatial data, represented primarily by remote sensing imagery, provides a comprehensive visual perspective on the Earth’s surface. However, interpreting such data remains challenging due to the diverse scales, shapes, and spatial layouts of geospatial objects~\cite{Wang_2025_CVPR_XLRSBench, gong2024crossearth}, as well as variations in illumination, sensor characteristics, and environmental conditions~\cite{an2025choice, wang2025hypersigma}.

Multimodal large language models (MLLMs) \cite{liu2023visual,zhu2023minigpt,alayrac2022flamingo,Dai2023InstructBLIPTG,bai2023qwen} have been introduced to enhance geospatial understanding through instruction tuning on remote sensing datasets \cite{kuckreja2024geochat,pang2025vhm,luo2024skysensegpt}. Although these models can handle basic remote sensing dialogues, they often fail in complex or unseen environments. This limitation mainly stems from relying solely on supervised fine-tuning (SFT), which leads the model to memorize linguistic patterns rather than develop genuine geospatial reasoning \cite{sft_memory_rl_generalize}.

\begin{figure}[t]
    \centering
    \includegraphics[width=\linewidth]{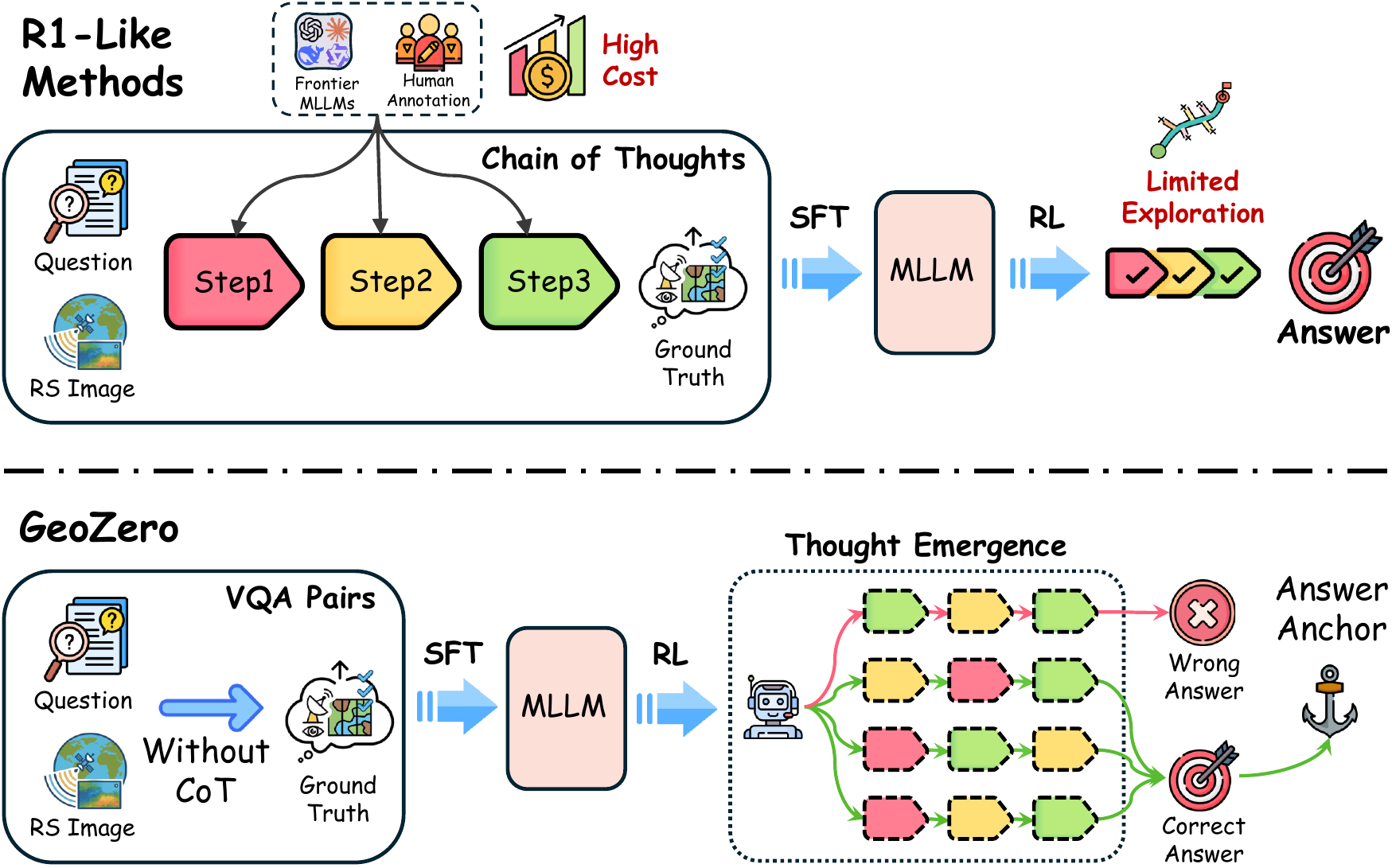}
    \caption{Compared with previous methods, GeoZero enables the emergence of reasoning on geospatial scenes without any CoT supervision while maintaining answer correctness.}
    \label{fig:motivation}
\end{figure}

Recently, advanced language models such as GPT-o1 \cite{gpt-o1} and DeepSeek-R1 \cite{deepseek-r1} have demonstrated explicit reasoning capabilities, characterized by performing deliberate thought before producing answers. Inspired by this progress, the remote sensing community has begun to explore MLLMs capable of “thinking” about geospatial scenes \cite{koksal2025tinyrs,scorers-r1}. Most existing approaches follow the DeepSeek-R1 training paradigm (see Figure \ref{fig:motivation}):
(1) constructing chain-of-thought (CoT) data for SFT to initialize the model with a basic “thinking–answering” response pattern, and
(2) applying reinforcement learning (RL), such as Group Relative Policy Optimization (GRPO), to further enhance its reasoning capability.
In these frameworks, the CoT annotations are typically generated by advanced general-purpose MLLMs (e.g., GPT-4o \cite{hurst2024gpt4o}) and subsequently verified or refined through manual inspection.

By examining the above process, we argue that, beyond the substantial annotation cost arising from the professional nature of remote sensing interpretation~\cite{dotav2}, training on manually revised reasoning paths inherently limits the model’s cognitive potential. In our view, manually refining CoTs inevitably introduces subjective biases that may constrain the model’s ultimate performance \cite{humanbias_emnlp,llm_bias,human_ai_act_bias}. To this end, a critical question naturally arises:  

\begin{framed}
     \textbf{\textit{Can multimodal large language models exhibit genuine and universal geospatial reasoning capabilities without any predefined chain-of-thought supervision?}}
\end{framed}

A preliminary analysis (see Section-\ref{sec:general_reason} in the supplementary material) shows that general instruction-tuned MLLMs exhibit extremely limited geospatial reasoning ability. To address this question, we present GeoZero, a MLLM capable of performing entirely internal reasoning on geospatial scenes without relying on any predefined CoT supervision. To achieve this, we follow the standard SFT–RL training paradigm commonly used by multimodal reasoning models~\cite{koksal2025tinyrs,vision-r1} and construct two remote sensing vision–language datasets: GeoZero-Instruct and GeoZero-Hard (both containing only question–answer pairs without any CoT data).
In the SFT stage, GeoZero-Instruct adapts general-purpose MLLMs to geospatial scenarios, enabling the acquisition of fundamental domain knowledge.
In the RL stage, since simple questions can often be answered without genuine reasoning, GeoZero-Hard is employed to promote deep reasoning by exposing the model to the most challenging examples.
Furthermore, we introduce an enhanced RL optimization framework, Answer-Anchored GRPO (A$^2$GRPO), where ``anchored'' denotes the joint refinement of reasoning quality and answer correctness based on the model’s predicted answers, thereby encouraging diverse yet accurate reasoning.

The main contributions of this paper are threefold:

(1) We propose GeoZero, a MLLM for geospatial understanding. It is capable of performing fully emergent reasoning on remote sensing imagery before answering new questions. To the best of our knowledge, this is the first MLLM in the remote sensing community that can reason without any predefined CoT supervision.

(2) We construct two remote sensing vision-language datasets: GeoZero-Instruct and GeoZero-Hard, and design a new optimization algorithm, A$^2$GRPO, to encourage MLLMs to generate diverse yet accurate reasoning during reinforcement learning. These designs enable the model to develop deep geospatial cognition and to think from scratch over complex geospatial scenes.

(3) We conduct extensive experiments on various remote sensing vision–language benchmarks, where GeoZero not only achieves competitive performance compared with existing methods but also exhibits emergent deep reasoning. Before answering, GeoZero generates reasoning trajectories from its internal cognition, demonstrating adaptive reasoning capabilities across diverse geospatial tasks.

\section{Related Work}

\subsection{MLLMs on Geospatial Scenes}

Benefiting from the rapid progress of general-purpose MLLMs \cite{gpt3, touvron2023llama, liu2024deepseek, Dai2023InstructBLIPTG, liu2023visual, zhu2023minigpt}, numerous geospatial-oriented MLLMs have recently been proposed \cite{hu2025rsgpt, kuckreja2024geochat, zhan2025skyeyegpt, liu2024rsunivlm, wang2024ringmogpt, wang2025geollavak, soni2024earthdial}. Due to the scarcity of textual annotations in the remote sensing community, early efforts relied on manual captioning \cite{hu2025rsgpt} or label-to-text conversion strategies \cite{zhang2024earthgpt} to construct image–text pairs for model training.
Unlike these manually generated datasets, later methods employed frontier MLLMs to perform automatic textual annotation \cite{kuckreja2024geochat,li2025lhrs,pang2025vhm,luo2024skysensegpt}, significantly improving data generation efficiency. Recent works have also addressed domain-specific challenges such as large-size \cite{wang2025geollavak, lrsvqa}, multi-temporal \cite{irvin2025teochat, li2024unirs}, and multi-sensor \cite{soni2024earthdial, shu2025earthmind} imagery. Beyond image–text dialogue systems, several MLLMs have been developed for vision-centric interpretation tasks \cite{zhou2024geoground, si2025spex, shabbir2025geopixel, ou2025geopix, zheng2025instructsam}. However, none of these models are capable of generating explicit reasoning chains before producing an answer. Their knowledge primarily stems from memorized textual patterns acquired through SFT, which limits their ability to generalize to complex and unseen geospatial scenarios.

\subsection{Large Multimodal Reasoning Models}

 The reasoning capability of large language models (LLMs) \cite{gpt-o1, deepseek-r1} has been greatly enhanced through post-training \cite{post-training}, achieving impressive performance on complex logical tasks such as mathematical problem solving \cite{yang2025qwen3} and science  exploration \cite{llmchemistry}. Especially, encouraged by the success of DeepSeek-R1 \cite{deepseek-r1}, numerous approaches have adopted GRPO to enhance the reasoning capability of MLLMs \cite{vision-r1, peng2025lmm, mm-eureka, visual-rft, vlm-r1, r1-vl}. In the remote sensing domain, several preliminary explorations have also emerged \cite{koksal2025tinyrs, geo-r1, geovlm-r1, scorers-r1, rsthinker, segearth-r1, remotereasoner, uav-vl-r1}. TinyRS-R1 \cite{koksal2025tinyrs} conducts GRPO on models fine-tuned with its constructed reasoning dataset, where the CoT is generated by GPT-4.1-mini through prompting with images, question–answer pairs, and system messages. ScoreRS \cite{scorers-r1} and Geo-R1 \cite{geo-r1} further improve reasoning performance on visual grounding and few-shot referring expression tasks. Compared with the above methods, our model is capable of performing emergent multimodal reasoning across multiple remote sensing vision–language tasks. It generates thinking trajectories entirely from its own internal cognition, without relying on any CoT annotations, thereby enabling emergent reasoning in geospatial understanding.

 \begin{figure*}[t]
    \centering
    \includegraphics[width=\linewidth]{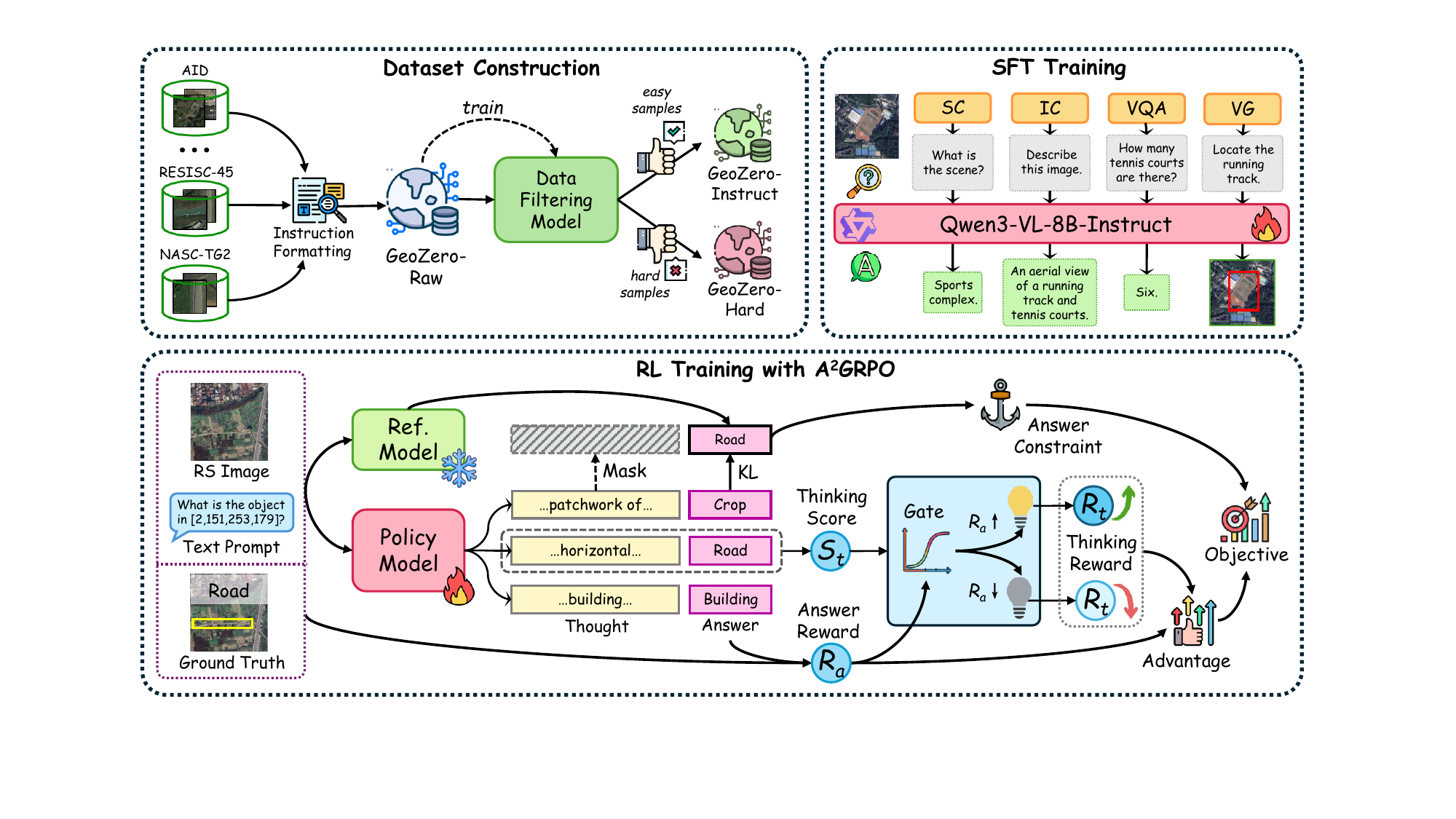}
    \caption{
    Overall framework of GeoZero. We first construct two datasets, GeoZero-Instruct and GeoZero-Hard, for SFT and RL, respectively. 
    The SFT stage leverages remote sensing instruct-following data without CoT to help MLLMs acquire basic geospatial knowledge, while the subsequent RL stage with A$^2$GRPO further enhances the reasoning ability, enabling the model to think from scratch.
    }
    \label{fig:Overall_framework}
\end{figure*}

\section{Method}

In this section, we present the development of GeoZero, including its overall training paradigm, the construction of GeoZero-Instruct and GeoZero-Hard, and the formulation of the RL algorithm.

\subsection{Overall Training Paradigm}

Figure~\ref{fig:Overall_framework} illustrates the overall framework of GeoZero, which employs a standard SFT–RL training paradigm. Following common practices in remote-sensing MLLMs~\cite{zhan2025skyeyegpt,zhang2024earthgpt}, we first conduct an SFT without CoT annotations on a constructed large-scale instruction-tuning dataset, GeoZero-Instruct, to equip the model with a preliminary understanding of geospatial scenes. In the subsequent RL stage, we construct a dedicated hard-sample pool, GeoZero-Hard, exposing the model to more challenging cases to stimulate deep reasoning. This two-stage framework allows the model to first acquire general geospatial knowledge and then stimulate its reasoning ability on difficult examples.

While our goal is to maximize the depth of reasoning in MLLMs, such reasoning must also remain grounded in correctness.
To this end, we propose A$^2$GRPO, an enhanced reinforcement learning framework that integrates the Answer-Modulated Thinking Reward (AMTR) and task-specific answer reward.
AMTR encourages the generation of high-quality reasoning chains while maintaining answer correctness, whereas the answer rewards further adapt MLLMs to various geospatial tasks.
Beyond the reward design, we also refine the RL optimization objective to promote diverse reasoning exploration.
Finally, we introduce a specialized system prompt during the RL stage to better guide the model’s reasoning behavior.

Next, we elaborate on the technical details of the key components introduced above.

\begin{table}[t]
\centering
\caption{Composition of the GeoZero-Raw dataset. ``Others'' include counting, honesty evaluation, and multi-turn dialogue samples without explicit task identifiers.}
\label{tab:GeoZero_raw}
\resizebox{\linewidth}{!}{
\begin{tabular}{llrc}
\toprule
Task Type & Dataset & \#Samples & Subtotal \\
\midrule
\multirow{7}{*}{Scene Classification}
 & VHM-Instruct \cite{pang2025vhm} & 9,050  & \multirow{7}{*}{110,649} \\
 & AID-train \cite{aid}                  & 8,000  &  \\
 & RESISC45-train \cite{nwpu}            & 25,200 &  \\
 & NASC-TG2-train \cite{nasc}            & 16,000 &  \\
 & WHU-RS19-train \cite{whurs19}         & 799    &  \\
 & EuroSAT-train \cite{eurosat}          & 21,600 &  \\
 & fMoW-RGB-train \cite{fmow}            & 30,000 &  \\
\midrule
\multirow{4}{*}{Visual Grounding}
 & VHM-Instruct \cite{pang2025vhm} & 53,982 & \multirow{4}{*}{126,620} \\
 & RSVG-train \cite{rsvg}                & 5,505  &  \\
 & DIOR-RSVG-train \cite{dior-rsvg}      & 30,820 &  \\
 & VRS-train \cite{li2024vrsbench}       & 36,313 &  \\
\midrule
\multirow{4}{*}{Visual Question Answering}
 & VHM-Instruct \cite{pang2025vhm} & 10,000 & \multirow{4}{*}{243,041} \\
 & RSVQA-HR-train \cite{rsvqa}           & 80,000 &  \\
 & RSVQA-LR-train \cite{rsvqa}           & 67,228 &  \\
 & VRS-train \cite{li2024vrsbench}       & 85,813 &  \\
\midrule
\multirow{2}{*}{Image Captioning}
 & SkyEye968k \cite{zhan2025skyeyegpt} & 153,375 & \multirow{2}{*}{173,639} \\
 & VRS-train \cite{li2024vrsbench}           & 20,264  &  \\
\midrule
\multirow{1}{*}{Others}
 & VHM-Instruct \cite{pang2025vhm} & 100,800 & \multirow{1}{*}{100,800} \\
\midrule
Total & - & - & 754,749 \\
\bottomrule
\end{tabular}
}
\end{table}

\subsection{Dataset Construction}

To establish the foundation for model training, we first aggregate publicly available remote sensing datasets into a unified collection, referred to as GeoZero-Raw. Based on this collection, we further derive two specialized subsets: GeoZero-Instruct for SFT and GeoZero-Hard for RL.

\subsubsection{GeoZero-Raw}
\label{sec:GeoZero_raw}

We mainly focus on four representative remote sensing vision–language tasks: scene classification (SC), visual grounding (VG), visual question answering (VQA), and image captioning (IC).
For each task, we collect samples from multiple classical datasets.
For instance, the SC task includes data from the training sets of AID~\cite{aid}, RESISC-45~\cite{nwpu}, and NASC-TG2~\cite{nasc}, among others.
The resulting GeoZero-Raw dataset spans diverse resolutions, sensor modalities, and scene categories, which helps enhance the generalization ability of MLLMs for remote sensing understanding.
A detailed composition of the dataset is summarized in Table~\ref{tab:GeoZero_raw}.

\noindent\textbf{Instruction Formatting.} After dataset aggregation, each sample is converted into an instruction-following format using a unified conversion scheme (see Section-\ref{sec:instruct_format} in the supplementary material). It should be noted that during dataset aggregation, some samples may share the same imagery from different datasets; however, since their textual instructions are regenerated under these rules, they are treated as distinct entries. In total, the resulting GeoZero-Raw dataset contains about 750K samples.

\subsubsection{GeoZero-Instruct and GeoZero-Hard}

In our consideration, to promote the emergence of genuine reasoning, the data used for RL should ideally be kept separate from the SFT stage.
Since MLLMs may, like humans, respond impulsively to straightforward or visually intuitive scenes, it becomes necessary to filter out such easy samples.
To this end, we construct a dedicated hard-sample pool, GeoZero-Hard, for the RL stage, ensuring that the model is exposed to the most challenging cases during training, thereby encouraging it to engage in genuine reasoning when confronted with complex geospatial tasks.

We design a three-step process to build these datasets:
(1) train the MLLM on GeoZero-Raw to obtain a data filtering model (DFM);
(2) select difficult samples to construct GeoZero-Hard; and
(3) perform data filtering based on GeoZero-Hard to produce GeoZero-Instruct. 
Intuitively, according to the maximum likelihood principle, samples mispredicted in a single inference pass are more likely to be difficult. Therefore, to ensure both the difficulty of GeoZero-Hard and the efficiency of sample selection, we adopt a coarse-to-fine two-stage filtering strategy. First, all samples in GeoZero-Raw are fed into the DFM for one inference pass, and only those incorrectly predicted are retained as potential hard-sample candidates. Next, these candidates are re-evaluated three times using the same model with different random seeds. Finally, we rank all candidates by their average error rate and select the top portion as the final hard samples. To maintain task balance, an equal number of samples are drawn from each of the four primary tasks. The selected samples collectively constitute the hard-sample pool, referred to as GeoZero-Hard. More details on the DFM and the hard-sample filtering procedure are provided in Section-\ref{sec:produce_hard} of the supplementary material.

Notably, to avoid potential image overlap within GeoZero-Raw (as discussed in Section-\ref{sec:GeoZero_raw}), we remove all samples sharing the same image with GeoZero-Hard. The remaining samples form GeoZero-Instruct. As a result, GeoZero-Hard and GeoZero-Instruct contain approximately 20K and 610K samples, respectively.

\subsection{Answer-Anchored Group Relative Policy Optimization}

As mentioned above, our RL training builds upon A$^2$GRPO, introducing two key enhancements: reward function and optimization objective.

\subsubsection{Reward Function}

Since the prompt template used in our RL stage is simplified (details introduced later), A$^2$GRPO differs from the original GRPO by coupling the thinking–answering format into a single constraint: responses with incorrect structures receive zero reward.
Otherwise, the reward is composed of two terms: thinking and answer rewards, measuring reasoning quality and answer correctness, respectively.

\noindent\textbf{Answer-Modulated Thinking Reward.} AMTR begins with a thinking-quality score \( s_t \), which comprehensively evaluates multiple aspects of the reasoning process, such as trajectory length, semantic diversity, and sentence redundancy. More details on \( s_t \) are provided in the supplementary material (Section-\ref{sec:think_quality_score}).

To make better use of this score, we further constrain it by the answer quality. Specifically, the thinking-quality score \( s_t \) is used only when the answer is sufficiently accurate. Moreover, once the answer becomes reliable, the contribution of reasoning is scaled proportionally to the answer quality. Accordingly, the AMTR \( r_t \) is formulated under dual answer constraints:

\begin{equation}
r_t = \mathcal{G}(r_a) \cdot r_a \cdot s_t,
\end{equation}
where
\begin{equation}
\mathcal{G}(r_a) = \frac{1}{1 + e^{-k(r_a - \tau)}}.
\end{equation}
Here, we employ a gate mechanism conditioned on the answer quality. Specifically, $r_a \in [0,1]$ denotes the answer reward (defined later), $\tau$ is a threshold that the answer is expected to exceed, and $k$ controls the sharpness of the activation of the thinking reward. In addition, the multiplicative term $r_a$ further constrains the thinking reward according to the overall answer quality.

In summary, AMTR encourages the model to enhance reasoning quality while maintaining answer correctness, thereby promoting the emergence of meaningful thinking behaviors during the RL stage.

\noindent\textbf{Task-Specific Answer Reward.} To adapt MLLMs to remote sensing vision-language tasks, following existing practices~\cite{koksal2025tinyrs,geovlm-r1,uav-vl-r1}, we additionally design task-specific answer rewards \( r_a \) for SC, VG, VQA, and IC. For consistency in reward computation, \( r_a \) is expected to fall within the range of \([0,1]\). Moreover, \( r_a \) should preferably be continuous so that it can smoothly reflect the quality differences among various answers, thereby facilitating model optimization during RL training. To this end, we formulate dedicated reward functions for each task, and the detailed formulations are provided in the supplementary material (Section-\ref{sec:task_answer_reward}).

Finally, the overall reward of A$^2$GRPO is formulated as:
\begin{equation}
    r = r_a + \lambda \cdot r_t,
    \label{all_reward}
\end{equation}
where $\lambda$ balances the contribution of the thinking process relative to the overall response quality.

\subsubsection{Optimization Objective}

In the original GRPO framework~\cite{deepseek-r1}, the optimization objective depends on the logits of responses containing both the reasoning trajectory and the final answer. However, since our goal is to incentivize MLLMs to think from scratch, \ie, without exposure to any CoT data, the reference model (the MLLM after SFT) does not explicitly generate reasoning text. In this case, directly applying GRPO's original formulation may inadvertently constrain the reasoning process. To address this issue, we refine the optimization objective in A$^2$GRPO as follows:

\begin{equation}
\resizebox{\linewidth}{!}{$
\begin{aligned}
\mathcal{J}_{\text{A}^2\text{GRPO}}(\theta) 
&= 
\mathbb{E}_{\{o_i\}_{i=1}^G \sim \pi_{\theta_{\text{old}}}(q)} 
\Bigg\{
    \frac{1}{G} \sum_{i=1}^{G} 
    \frac{1}{|o_i|} \sum_{t=1}^{|o_i|}
    \Big[
        \min\big(
            \varphi_{i,t} A_i,\;
         \\   &\text{clip}(\varphi_{i,t}, 1-\varepsilon, 1+\varepsilon) A_i
        \big)
        - \beta\, m_{i,t}\,
        \big(
            \phi_{i,t} - \log \phi_{i,t} - 1
        \big)
    \Big]
\Bigg\},
\end{aligned}
$}
\label{eq:a2grpo}
\end{equation}
where 
\begin{equation}
\small
    \varphi_{i,t} = 
    \frac{\pi_{\theta}(o_{i,t}|q, o_{i,<t})}
         {\pi_{\theta_{\text{old}}}(o_{i,t}|q, o_{i,<t})},
    \quad
    \phi_{i,t} = 
    \frac{\pi_{\theta_{\text{ref}}}(o_{i,t}|q, o_{i,<t})}
         {\pi_{\theta}(o_{i,t}|q, o_{i,<t})}.
\label{eq:ratios}
\end{equation}

The definitions of \( G \), \( \varepsilon \), \( \beta \), and the advantage term \( A_i(r_1,\dots,r_G) \) follow the original GRPO~\cite{deepseek-r1}, where \( r_i \) denotes the overall reward (see Equation~(\ref{all_reward})) for each candidate response.
The key difference lies in the introduction of a binary mask \( M_i=\{m_{i,t}\}_{t=1}^{|o_i|} \), where \( m_{i,t}=1 \) if token \( t \) is located within the answer tags (as detailed later), and \( m_{i,t}=0 \) otherwise. This design ensures that the distributional alignment with the reference model is enforced only on the answer portion, thereby preventing the reasoning tokens from being overly regularized and encouraging MLLMs to freely explore diverse reasoning trajectories.

\subsubsection{Prompt template}

Despite these efforts, MLLMs may still fail to generate explicit reasoning texts. In our experiments, we observed that directly using the original GRPO template~\cite{deepseek-r1} (\textit{\textless think\textgreater reasoning process\textless/think\textgreater \textless answer\textgreater answer\textless/answer\textgreater}) often fails to activate reasoning behaviors, as tokens "\textit{\textless think\textgreater}" or "\textit{\textless}" rarely appear at the beginning of sentences for SFT. Consequently, the model struggles to initiate coherent reasoning from scratch. To alleviate this issue, we simplify the GRPO prompt template by removing corresponding constraints while retaining only the "\textit{\textless answer\textgreater\textless/answer\textgreater}" markers. More details of our system prompts are provided in Section-\ref{sec:prompt_template} of the supplementary material.

\section{Experiment}

In this section, we evaluate GeoZero through both quantitative and qualitative experiments. 
We first report results on four typical remote sensing vision-language tasks: scene classification (SC), visual grounding (VG), visual question answering (VQA), and image captioning (IC). 
We then conduct ablation studies to analyze the contribution of our datasets and RL algorithm, followed by an analysis of the correlation between thinking behaviors and answer correctness. 
Finally, we visualize representative reasoning examples to illustrate GeoZero's emergent geospatial reasoning.

\subsection{Experimental Settings}

We train GeoZero following a two-stage SFT–RL framework, with Qwen3-VL-8B-Instruct \cite{Qwen3-VL} serving as the base model. Both SFT and RL are trained for 1 epoch. By controlling the gradient accumulation steps, we set the global batch size to 64 and 48, respectively. The learning rates for SFT and RL are 1e-4 and 5e-6, respectively, with weight decays of 0.1 and 0.01. During both stages, the vision encoder is kept frozen, while the remaining modules are fine-tuned using LoRA \cite{lora} with a rank of 16 and $\alpha = 32$. We employ bfloat16 precision and DeepSpeed \cite{deepspeed} to reduce GPU memory consumption, and use FlashAttention \cite{dao2022flashattention} to accelerate computation. For A$^2$GRPO, the group number $G$ is set to 8, the temperature is 0.9, and $\beta$ in Equation (\ref{eq:a2grpo}) is 0.04. For AMTR, the gating parameters $k$ and $\tau$ are 12 and 0.5. The coefficient $\lambda$ in Equation~(\ref{all_reward}) is set to 0.3. All experiments are conducted on 8 NVIDIA A100 GPUs.

\subsection{Comparison to State-of-the-art MLLMs}

\begin{table}[t]
  \small
  \caption{Performance comparison (\%) of different MLLMs on scene classification tasks. The best and second-best results are highlighted in \textbf{bold} and \textcolor{blue}{blue}, respectively.}
  \centering
  \begin{tabular}{lcc}
  \toprule
  Method 
  & UCM 
  & AID\\
  \midrule
  Qwen3-VL-8B-Instruct \cite{Qwen3-VL}  & 75.71 & 71.40 \\
  GeoChat \cite{kuckreja2024geochat}     & 84.43 & 72.03 \\
  VHM \cite{pang2025vhm}                 & -    & \textcolor{blue}{91.70} \\
  ScoreRS \cite{scorers-r1}               & -     & 85.90  \\
  RingMo-Agent \cite{ringmoagent}        & \textcolor{blue}{88.00} & 91.67 \\
  TinyRS-R1 \cite{koksal2025tinyrs}      & -    & 90.20 \\
  \midrule
  GeoZero                     & \textbf{93.81} & \textbf{92.55} \\
  \bottomrule
  \end{tabular}
  \label{tab:cls_result}
\end{table}

\begin{table}[t]
  \small
  \caption{Performance comparison (\%) of different MLLMs on visual grounding tasks. The best and second-best results are highlighted in \textbf{bold} and \textcolor{blue}{blue}, respectively.}
  \centering
  \begin{tabular}{lcc}
  \toprule
  Method 
  & RSVG 
  & DIOR-RSVG \\
  \midrule
  Qwen3-VL-8B-Instruct \cite{Qwen3-VL}   & 26.41 & 51.51 \\
  GeoChat \cite{kuckreja2024geochat}      & 14.67 & 19.77 \\
  VHM \cite{pang2025vhm}                       &  -    & 56.17 \\
  ScoreRS-R1 \cite{scorers-r1}                 &  -    & 64.52 \\
  TinyRS-R1 \cite{koksal2025tinyrs}       & -    & 74.90 \\
  GeoGround \cite{zhou2024geoground}      & \textcolor{blue}{26.65} & \textbf{77.73} \\
  \midrule
  GeoZero                      & \textbf{37.16} & \textcolor{blue}{75.67} \\
  \bottomrule
  \end{tabular}
  \label{tab:vg_result}
\end{table}

\begin{table}[t]
  \small
  \caption{Performance comparison (\%) of different MLLMs on visual question answering tasks. The best and second-best results are highlighted in \textbf{bold} and \textcolor{blue}{blue}, respectively.}
  \centering
  \begin{tabular}{lcc}
  \toprule
  \multirow{2}*{Method} & \multicolumn{2}{c}{RSVQA-HR} \\
  \cmidrule{2-3}
  & Presence & Compare \\
  \midrule
  Qwen3-VL-8B-Instruct \cite{Qwen3-VL}   & 66.04 & 75.39 \\
  GeoChat \cite{kuckreja2024geochat}      & 58.45 & 83.19 \\
  VHM     \cite{pang2025vhm}          & 64.00 & \textcolor{blue}{83.50} \\
  TinyRS-R1  \cite{koksal2025tinyrs}        & \textcolor{blue}{68.60} & 73.50 \\
  \midrule
  GeoZero                      & \textbf{74.46} & \textbf{83.59} \\
  \bottomrule
  \end{tabular}
  \label{tab:vqa_result}
\end{table}

\begin{table}[t]
  \small
\caption{Performance comparison (\%) of different MLLMs on image captioning tasks. 
  The best and second-best results are highlighted in \textbf{bold} and \textcolor{blue}{blue}, respectively.}
\centering
\begin{tabular}{lccc}
\toprule
\multirow{2}*{Method} & \multicolumn{3}{c}{RSICD} \\
 \cmidrule{2-4}
 & BLEU-4 & CIDEr & METEOR \\
\midrule
GeoChat \cite{kuckreja2024geochat}     & -    & 12.39 & 13.48 \\
SkyEyeGPT \cite{zhan2025skyeyegpt}    & \textbf{59.99} & \textcolor{blue}{83.65} & \textcolor{blue}{35.35} \\
\midrule
GeoZero  & \textcolor{blue}{29.17} & \textbf{97.45} & \textbf{48.18} \\
\bottomrule
\end{tabular}
\label{tab:ic_result}
\end{table}

\noindent\textbf{Typical Remote Sensing Vision-Language Tasks.}
Tables~\ref{tab:cls_result}-\ref{tab:ic_result} summarize the performance of GeoZero against representative state-of-the-art remote sensing MLLMs on four typical tasks: SC, VG, VQA, and IC. For evaluation, SC and VQA are measured by accuracy. VG adopts Acc@0.5, where a prediction is considered correct if the IoU between the predicted box and the ground truth exceeds 0.5. For IC, we report standard captioning metrics, including BLEU-4, CIDEr, and METEOR.

For SC, we evaluate on two widely used benchmarks, UCM~\cite{ucm} and AID~\cite{aid}, following common evaluation practices in recent works~\cite{pang2025vhm,koksal2025tinyrs,scorers-r1}. As shown in Table~\ref{tab:cls_result}, GeoZero achieves superior performance on both datasets, outperforming both general MLLMs and recent remote sensing MLLMs. Notably, GeoZero yields a clear margin on UCM, indicating strong generalization to unseen scenes.

For VG, we report results on RSVG~\cite{rsvg} and DIOR-RSVG~\cite{dior-rsvg}. As shown in Table~\ref{tab:vg_result}, GeoZero delivers the best performance on RSVG and remains highly competitive on DIOR-RSVG, approaching the specialized grounding model GeoGround~\cite{zhou2024geoground}. These results demonstrate that our method effectively strengthens spatial localization and grounding capability.

For VQA, we adopt RSVQA-HR~\cite{rsvqa} for evaluation and report accuracy on its two question types (Presence and Compare). The results in Table~\ref{tab:vqa_result} show that GeoZero attains the highest accuracy on both question types, improving substantially over the general-purpose base model while remaining competitive against prior remote sensing MLLMs.

Finally, for IC, we evaluate on RSICD~\cite{rsicd}. As shown in Table~\ref{tab:ic_result}, GeoZero achieves the best CIDEr and METEOR among the compared methods, despite CIDEr not being explicitly optimized in the task-specific answer reward. This suggests that the proposed training framework can improve the overall quality of generated descriptions beyond the directly optimized components.

\subsection{Ablation Study}

To validate the effectiveness of the constructed datasets and the proposed RL algorithm, we conduct ablation studies on two representative tasks: SC and VG, using the UCM and RSVG datasets, respectively.

Notably, to further analyze the activation of reasoning in task-specific scenarios, in addition to the standard SFT and RL stages, we additionally introduce reinforcement fine-tuning (RFT) with A$^2$GRPO on each dataset (denoted as GeoZero+RFT). Specifically, \textbf{RFT is performed only on the training split of each dataset, with no overlap with the evaluation data.} The detailed configurations of RFT are provided in Section~\ref{sec:rft_settings} of the supplementary material.

\begin{table*}[t]
  \small
  \caption{Ablation on different training settings. {\tiny$\bullet$} and {\tiny$\circ$} indicate whether thinking texts are generated during inference ({\tiny$\bullet$}: with reasoning, {\tiny$\circ$}: without reasoning).}
  \centering
  \begin{tabular}{lccc|cc}
  \toprule
  Model & SFT & RL & RFT & UCM (\%) & RSVG (\%) \\
  \midrule
  Qwen3-VL-8B-Instruct  &  &  &  & 75.71 \,({\tiny$\circ$}) & 26.41 \,({\tiny$\circ$}) \\
  Qwen3-VL-8B-Instruct (RL only) &  & \checkmark &  & 76.90 \,({\tiny$\bullet$}) & 0.16 \,({\tiny$\bullet$}) \\
  Qwen3-VL-8B-Instruct (RL+RFT) &  & \checkmark & \checkmark & 76.19 \,({\tiny$\bullet$}) & 28.12 \,({\tiny$\bullet$}) \\
  Qwen3-VL-8B-Instruct (SFT only) & \checkmark &  &  & 94.52 \,({\tiny$\circ$}) & 47.76 \,({\tiny$\circ$}) \\
  Qwen3-VL-8B-Instruct (SFT+RFT) & \checkmark &  & \checkmark & 95.24 \,({\tiny$\circ$}) & 47.68 \,({\tiny$\circ$}) \\
  \midrule
  GeoZero & \checkmark & \checkmark &  & 93.81 \,({\tiny$\bullet$}) & 37.16 \,({\tiny$\bullet$}) \\
  GeoZero+RFT  & \checkmark & \checkmark & \checkmark & 95.48 \,({\tiny$\bullet$}) & 50.04 \,({\tiny$\bullet$}) \\
  \bottomrule
  \end{tabular}
  \label{tab:ablation_stage}
  
\end{table*}

\begin{table}[t]
  \scriptsize
  \caption{Ablation study on the dataset used during RL. ``Random'' means the data are randomly selected from GeoZero-Raw-val (see Section-\ref{sec:produce_hard} in the supplementary material), while the construction method of the SFT dataset is the same as GeoZero-Instruct. {\tiny$\bullet$} and {\tiny$\circ$} follow the same definitions as in Table~\ref{tab:ablation_stage}.
  }
  \centering
  \resizebox{\linewidth}{!}{
  \begin{tabular}{ll|cc}
  \toprule
  Model & RL Data & UCM (\%) & RSVG (\%)\\
  \midrule
  GeoZero & Random           & 94.05 \,({\tiny$\circ$}) & 18.34 \,({\tiny$\circ$}) \\
  GeoZero & GeoZero-Hard  & 93.81 \,({\tiny$\bullet$}) & 37.16 \,({\tiny$\bullet$}) \\
  \midrule
  GeoZero+RFT  & Random           & 95.24 \,({\tiny$\circ$}) & 46.70 \,({\tiny$\bullet$}) \\
  GeoZero+RFT  & GeoZero-Hard  & 95.48 \,({\tiny$\bullet$}) & 50.04 \,({\tiny$\bullet$}) \\
  \midrule
  \end{tabular}
  }
  \label{tab:ablation_data}
  
\end{table}

\begin{table}[t]
  \scriptsize
  \caption{Ablation on different components of A$^2$GRPO.   {\tiny$\bullet$} and {\tiny$\circ$} follow the same definitions as in Table~\ref{tab:ablation_stage}.
  }
  \centering
  \resizebox{\linewidth}{!}{
  \begin{tabular}{lcc|cc}
  \toprule
  Model & AMTR & Thinking Mask & UCM (\%) & RSVG (\%)\\
  \midrule
  GeoZero &     &   &  91.67 \,({\tiny$\circ$}) &   26.49 \,({\tiny$\circ$}) \\
  GeoZero  &  & \checkmark & 93.10 \,({\tiny$\circ$}) &  28.28 \,({\tiny$\circ$})  \\
  GeoZero  & \checkmark &  & 93.57 \,({\tiny$\circ$})  & 26.57 \,({\tiny$\circ$})     \\
  GeoZero   & \checkmark & \checkmark & 93.81 \,({\tiny$\bullet$}) & 37.16 \,({\tiny$\bullet$})\\
  \midrule
  GeoZero+RFT &     &   &  93.81 \,({\tiny$\circ$}) &   33.01 \,({\tiny$\circ$})   \\
  GeoZero+RFT  &  & \checkmark & 95.24 \,({\tiny$\circ$}) & 49.80 \,({\tiny$\circ$}) \\
  GeoZero+RFT  & \checkmark &     & 94.76 \,({\tiny$\bullet$}) & 47.03 \,({\tiny$\bullet$}) \\
  GeoZero+RFT  & \checkmark & \checkmark & 95.48 \,({\tiny$\bullet$}) & 50.04 \,({\tiny$\bullet$}) \\
  \bottomrule
  \end{tabular}
  }
  \label{tab:ablation_a2grpo}
  
\end{table}

\noindent\textbf{Training Paradigm.}
We first examine the impact of different training strategies, as summarized in Table~\ref{tab:ablation_stage}. Direct RL training (RL only) can trigger reasoning generation ({\tiny$\bullet$}). However, without SFT, the model lacks fundamental geospatial knowledge and performs poorly on challenging benchmarks (e.g., RSVG). Although RFT can partially recover accuracy, it still falls short of the SFT-RL pipeline trained with GeoZero-Hard. In contrast, SFT-only and SFT+RFT achieve strong accuracy but fail to activate reasoning ({\tiny$\circ$}), likely due to the absence of CoT supervision in SFT and the limited scale and diversity of downstream training data (compared with GeoZero-Hard) in RFT, which provide insufficient optimization signals to induce reasoning. By incorporating the constructed hard-sample pool, GeoZero-Hard, into the SFT-RL paradigm (GeoZero w/o RFT), the model maintains high accuracy while consistently activating reasoning, indicating its ability to think from scratch. Further applying dataset-specific RFT (GeoZero w/ RFT) improves both task accuracy and reasoning consistency, validating the effectiveness of our training paradigm.

\noindent\textbf{GeoZero-Hard.} We evaluate the impact of the constructed datasets, as shown in Table \ref{tab:ablation_data}. It can be observed that when RL training uses randomly selected data instead of hard samples, although the model performs well on the UCM dataset, it fails to generate reasoning texts, and the performance significantly decreases on the RSVG dataset. This suggests that the model struggles with reasoning on more challenging scenes when trained with random samples. After RFT, the model reactivates reasoning on the more challenging RSVG dataset, yet its performance remains to be further improved. These results highlight the importance of using hard samples during RL training to help the model engage in genuine reasoning from scratch.

\noindent\textbf{A$^2$GRPO.} We further evaluate the effectiveness of the proposed A$^2$GRPO algorithm by examining the influence of the AMTR and the thinking mask. As shown in Table~\ref{tab:ablation_a2grpo}, without answer modulation (\ie, $r_t = s_t$), the emergence of reasoning becomes independent of the answer reward and is further constrained by factors such as KL penalties and redundancy suppression.  In this case, without CoT-based cold-start training, the model tends to converge to a risk-averse strategy: simply outputting answers without generating reasoning, so as to maintain stability. When AMTR is applied but the thinking mask is removed, additional constraints hinder the diversity of reasoning generation, thereby degrading model performance, particularly in reasoning-intensive scenarios (e.g., the RSVG dataset). In contrast, A$^2$GRPO encourages constructive reasoning while removing unintended constraints on the thinking process, thereby expanding the reasoning exploration space. As a result, our method effectively activates reasoning while maintaining high answer accuracy, ultimately enhancing the model’s emergent reasoning capability.

In summary, the results in Tables~\ref{tab:ablation_stage}–\ref{tab:ablation_a2grpo} demonstrate that, by leveraging hard samples and proper RL strategies (\ie, AMTR and the thinking mask), we successfully trigger the emergence of thinking in MLLMs on geospatial scenes, a capability that does not appear after SFT.

\subsection{Visualization}

\begin{figure}[t]
    \centering
    \subfigure[]{
        \includegraphics[width=0.85\linewidth]{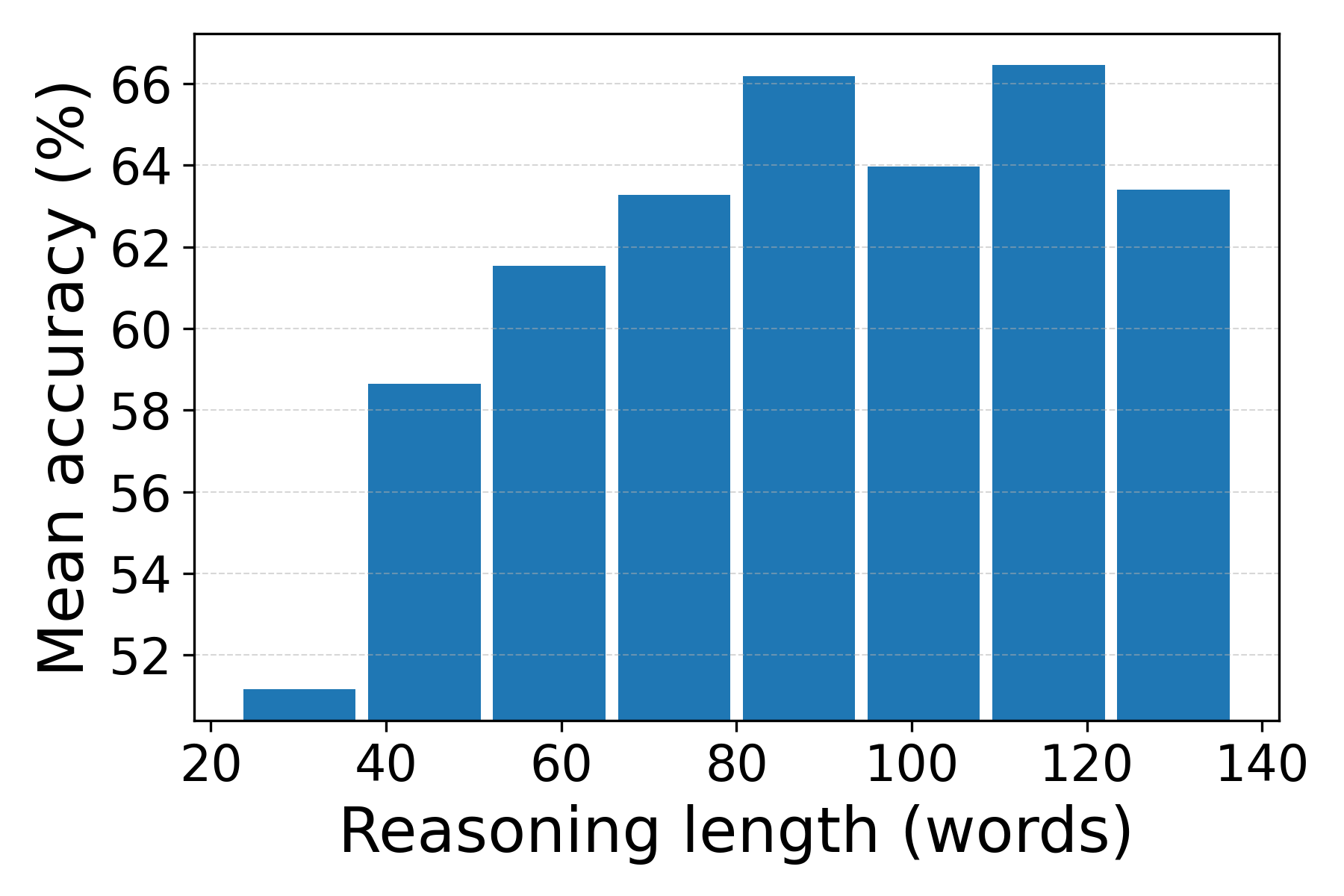}
    }
    \hfill
    \subfigure[]{
        \includegraphics[width=0.85\linewidth]{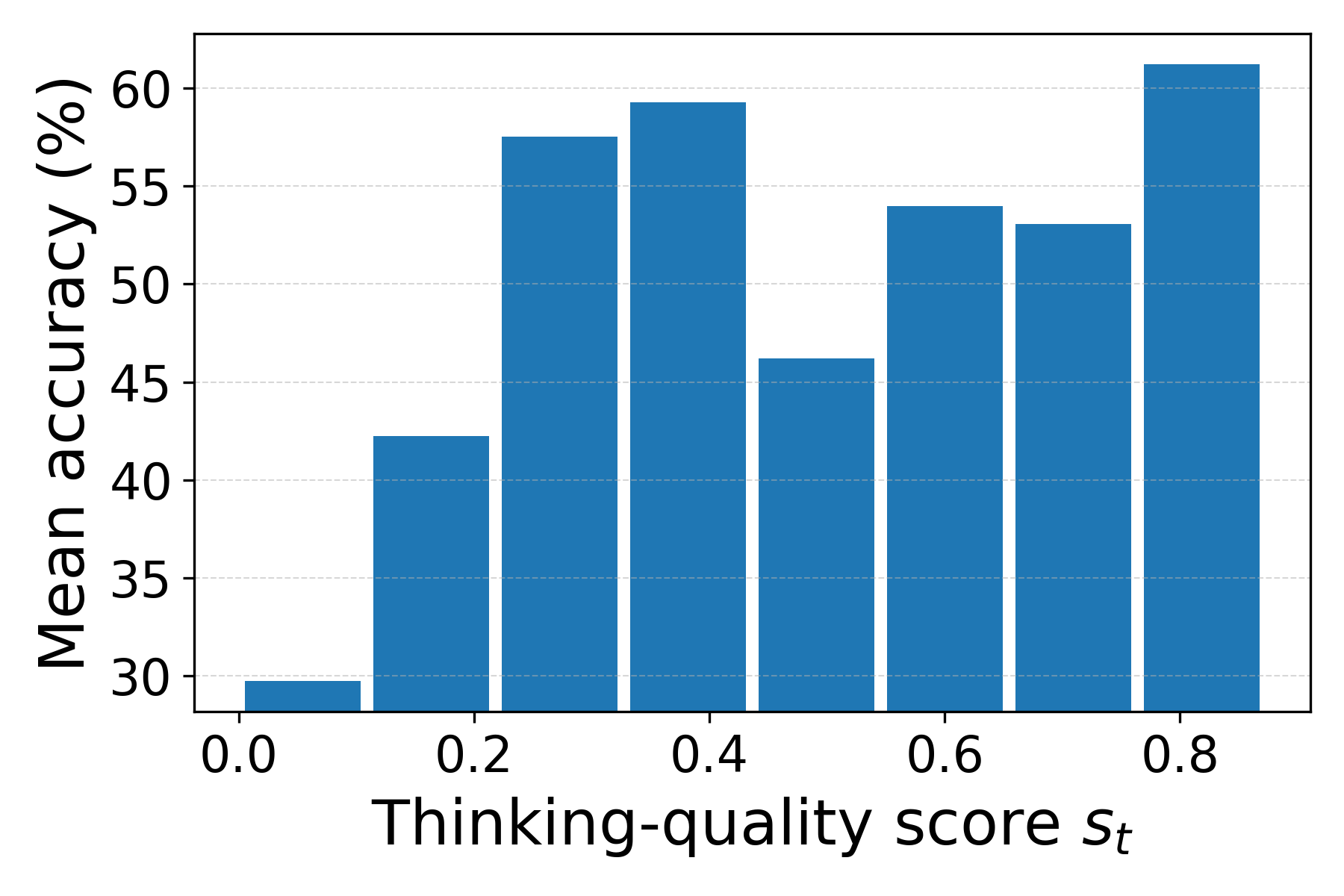}
    }
    \caption{Relationship between thinking and answering, where the statistics are computed from GeoZero’s prediction results on the DIOR-RSVG test set. (Best view in zoom.)}
    \label{fig:think_vs_answer}
    
\end{figure}

\noindent\textbf{Thinking v.s. Answering.} 
To further investigate how reasoning relates to final predictions, we analyze the relationship between answer accuracy and both reasoning length and reasoning quality score $s_t$. Specifically, all samples are grouped based on their reasoning length and quality score, and the average accuracy within each group is computed. As shown in Figure~\ref{fig:think_vs_answer}(a), accuracy increases with longer reasoning chains, indicating that extended reasoning generally benefits answer correctness. However, when the reasoning becomes excessively long, \ie, overthinking, the performance slightly drops. Similarly, Figure~\ref{fig:think_vs_answer}(b) reveals a positive correlation between $s_t$ and answer accuracy, indicating that samples with higher reasoning quality scores tend to achieve higher accuracy. Overall, these results suggest a strong association between reasoning quality and task performance, supporting our design choice of encouraging explicit reasoning during RL. More fine-grained analyses are provided in Section-\ref{sec:analysis_reason} of the supplementary material.

\begin{figure}[t]
    \centering
    \includegraphics[width=\linewidth]{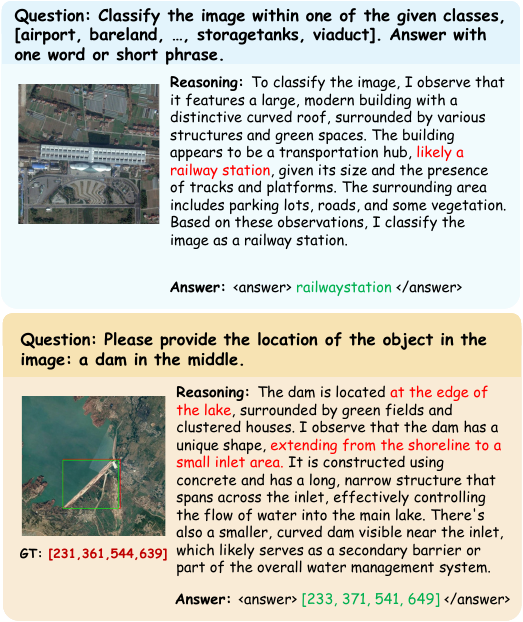}
    \caption{Visualization of reasoning processes generated by GeoZero on different tasks. Examples are drawn from the test sets of AID and DIOR-RSVG. \textcolor{red}{Red} text highlights the model's reasoning steps leading to the final answers. GT: ground truth.}
    \label{fig:reason_vis}
    
\end{figure}

\noindent\textbf{Qualitative Reasoning Results.}  
Finally, to gain an intuitive understanding of the model’s reasoning process, we qualitatively visualize several examples of GeoZero’s predictions, as shown in Figure~\ref{fig:reason_vis}. It can be observed that before producing the final answer, GeoZero explicitly articulates its internal reasoning by describing key visual attributes such as object shape, color, spatial layout, quantity, and contextual relationships. These examples demonstrate that GeoZero is capable of generating coherent and interpretable reasoning trajectories, showcasing its ability to perform emergent reasoning over complex geospatial scenes, while also highlighting its potential for building transparent and explainable AI systems. Additional reasoning examples are shown in Section-\ref{sec:visual_examples} of the supplementary material.

\section{Conclusion}

In this work, we present GeoZero, the first MLLM capable of performing emergent reasoning on geospatial scenes from scratch. To encourage deep and reliable reasoning while maintaining answer accuracy, we construct two datasets: GeoZero-Instruct for SFT and GeoZero-Hard for RL. We also propose a novel RL framework with tailored reward functions and optimization objectives. Through these designs, GeoZero successfully activates reasoning ability without any cold-start CoT supervision. Extensive experiments demonstrate that our model exhibits universal reasoning across multiple remote sensing vision–language tasks. Overall, our approach not only reduces annotation costs but also enhances the cognitive capability of MLLMs, offering new insights toward general geospatial AI.

\bibliography{egbib}
\bibliographystyle{IEEEtran}

\appendices


\section{Overview}

This supplementary material provides further details for the proposed model: GeoZero, as well as the developed datasets: GeoZero-Raw and GeoZero-Hard. These details were omitted from the main paper due to space constraints.

The supplementary material is organized as follows:

\begin{itemize}
    \item Section~\ref{sec:general_reason}: Evaluating whether general MLLMs can perform geospatial reasoning.
    \item Section~\ref{sec:instruct_format}: Implementation details of instruction formatting for constructing GeoZero-Raw.  
    \item Section~\ref{sec:produce_hard}: Detailed data filtering and selection process for producing GeoZero-Hard.  
    \item Section~\ref{sec:prompt_template}: System prompt design for GeoZero-Hard.  
    \item Section~\ref{sec:think_quality_score}: Calculation process of the thinking-quality score~$s_t$.  
    \item Section~\ref{sec:task_answer_reward}: Calculation of the task-specific answer reward~$r_a$ for different tasks.  
    \item Section~\ref{sec:rft_settings}: Specific reinforcement fine-tuning configurations across multiple datasets for GeoZero. 
    \item Section~\ref{sec:analysis_reason}: Further analysis of reasoning behavior.  
    \item Section~\ref{sec:visual_examples}: Additional qualitative prediction examples of GeoZero.  
    \item Section~\ref{sec:datasheet}: Datasheets for GeoZero-Raw.  
\end{itemize}

\section{Can General MLLMs Perform Geospatial Reasoning?}
\label{sec:general_reason}

The motivation of GeoZero is to enable multimodal large language models (MLLMs) to acquire geospatial reasoning abilities without relying on pre-defined chain-of-thought supervision, but instead to stimulate such reasoning through reinforcement learning strategies. However, since existing instruction-tuned MLLMs may already exhibit a certain degree of reasoning capability, it is important to verify whether the effectiveness of our method stems from the proposed framework itself rather than from the inherent reasoning ability of the base model.

To this end, we conduct a diagnostic experiment using the base model of GeoZero, Qwen3-VL-8B-Instruct. To ensure fairness, we align the evaluation configuration with that used for GeoZero in the main paper, where the model is guided by the same system prompt employed in the reinforcement learning stage, specifically the Type1 version without illustrative examples (see Figure~\ref{fig:system_prompt}). We evaluate the model on two remote sensing vision-language datasets, UCM~\cite{ucm} and RSVG~\cite{rsvg}, where the model is instructed to produce an explicit reasoning process before outputting the final answer.

To measure whether the model performs explicit reasoning, we adopt a simple yet permissive criterion: a sample is considered as “reasoned” if there exist any texts appearing before the ``\textit{\textless answer\textgreater}'' tag in the model’s output. Based on this rule, we calculate the proportion of samples exhibiting reasoning behavior (referred to as the "thinking activation rate"), and summarize the results in Table~\ref{tab:ucm_rsvg_think_or_not}.

\begin{table}[t]
  \centering
    \caption{Accuracy and thinking activation rate (\%) of different models on the UCM and RSVG datasets. GeoZero uses the system prompt (SP) by default. w/SP: with the system prompt (Type1) shown in Figure~\ref{fig:system_prompt}. Acc.: accuracy. TAR: thinking activation rate.}
     \newcommand{\tabincell}[2]{\begin{tabular}{@{}#1@{}}#2\end{tabular}}
  \resizebox{\linewidth}{!}{
  \begin{tabular}{l|cc|cc}
  \toprule
  \multirow{2}{*}{Model} & 
  \multicolumn{2}{c|}{UCM \cite{ucm}} & 
  \multicolumn{2}{c}{RSVG \cite{rsvg}} \\
  \cmidrule(lr){2-3} \cmidrule(lr){4-5}
   & Acc. (\%) & TAR (\%) & Acc. (\%) & TAR (\%) \\
   \midrule
   \tabincell{c}{Qwen3-VL-8B-Instruct \\(w/o SP)} &  75.71  & 0.00  & 26.41 & 0.00 \\
  \tabincell{c}{Qwen3-VL-8B-Instruct \\(w/ SP)} & 77.14 & \tabincell{c}{53.10\\(80.72\% corrected)\\(223/420)} & 24.04 & \tabincell{c}{0.08\\(1/1227)} \\
  \midrule
   GeoZero & 93.81 & \tabincell{c}{99.52\\(94.26\% corrected)\\(418/420)} & 37.16 &  \tabincell{c}{100 \\(1227/1227)} \\
  \bottomrule
  \end{tabular}
  }
  \label{tab:ucm_rsvg_think_or_not}
\end{table}

The results indicate that directly applying system prompts cannot improve the accuracy of general MLLM on both datasets (75.71\% $\rightarrow$ 77.14\%, 26.41\% $\rightarrow$ 24.04\%). On UCM, the model performs explicit reasoning on 53.10\% of the samples, achieving an accuracy of 80.72\% on these samples, which is higher than the overall accuracy of 77.14\%. This suggests that general-purpose MLLMs are capable of performing reasoning in certain geospatial contexts, and such reasoning indeed enhances performance.

However, on the more challenging RSVG dataset, even with reasoning prompts, the model almost never engages in explicit reasoning, directly outputting answers without any preceding reasoning process. Among the 1,227 test samples, only one contained non-empty reasoning content before the ``\textit{\textless answer\textgreater}'' tag. This observation highlights that \textbf{\textit{the geospatial reasoning capability of general MLLMs remains extremely limited.}}

In contrast, our GeoZero framework, consistently produces explicit reasoning traces while improving accuracy. On the UCM dataset, the reasoning activation rate is slightly below 100\%, likely because some samples are too simple to require reasoning. On the more complex RSVG dataset, GeoZero demonstrates full activation of geospatial reasoning, reaching a 100\% reasoning rate. \textbf{\textit{These findings further validate the rationale behind our research motivation.}}

\begin{figure*}
    \centering
    \includegraphics[width=0.8\linewidth]{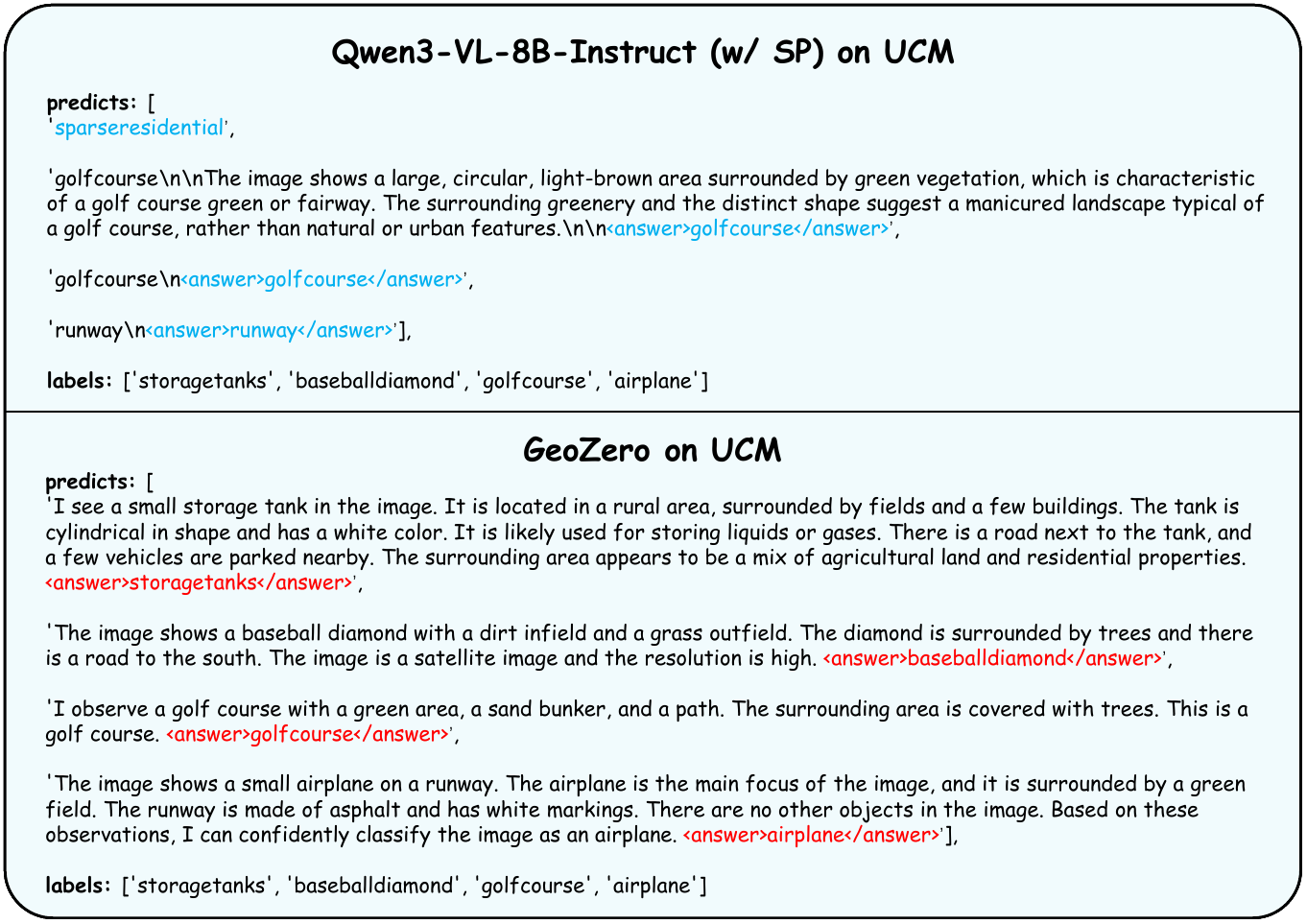}
    \caption{Reasoning outputs from different models on the UCM dataset for one iteration (batch size = 4).}
    \label{fig:reason_examples_ucm}
\end{figure*}

\begin{figure*}
    \centering
    \includegraphics[width=0.8\linewidth]{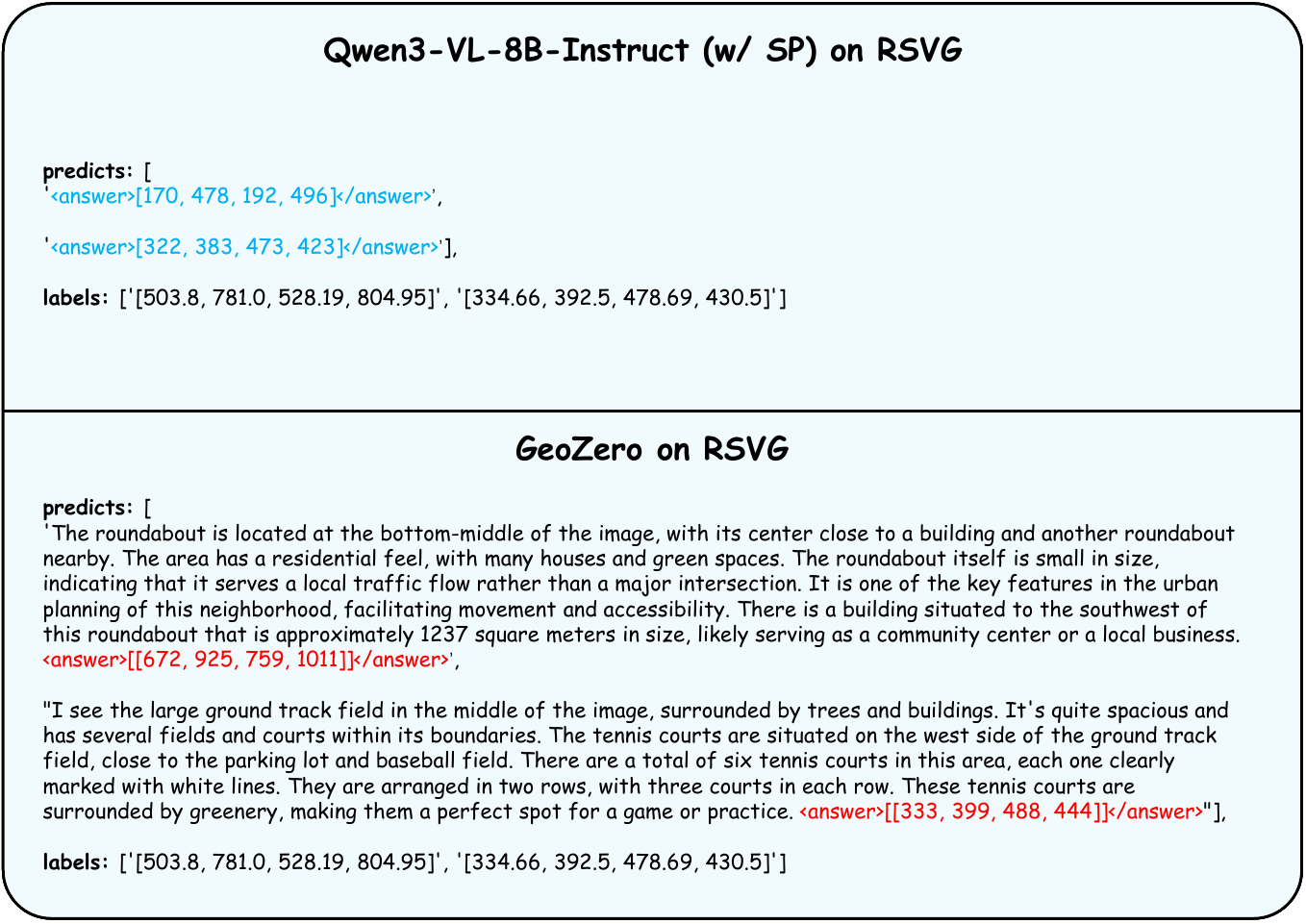}
    \caption{Reasoning outputs from different models on the RSVG dataset for one iteration (batch size = 2).}
    \label{fig:reason_examples_rsvg}
\end{figure*}

Figure~\ref{lst:think_activation} presents the Python code used to detect explicit reasoning traces in model outputs and compute their corresponding activation rate. Representative reasoning examples for different models on the UCM and RSVG datasets have been shown in Figure~\ref{fig:reason_examples_ucm}–\ref{fig:reason_examples_rsvg}.

\begin{figure}[t]
\centering
\setlength{\abovecaptionskip}{2pt}
\setlength{\belowcaptionskip}{2pt}

\begin{lstlisting}[language=Python]
def calc_thinking_ratio(results):
     total = len(results)
     triggered = 0

     for item in results:
         raw = item.get("raw_output", "")

         if "<answer>" not in raw:
             continue

         thought = raw.split("<answer>")[0].strip()
         if len(thought) > 0:
             triggered += 1

     ratio = triggered / total if total > 0 else 0

     return ratio
\end{lstlisting}

\caption{Example code for evaluating reasoning activation.}
\label{lst:think_activation}
\end{figure}

\section{Instruction Formatting of GeoZero-Raw}
\label{sec:instruct_format}

\begin{figure*}
    \centering
    \includegraphics[width=\linewidth]{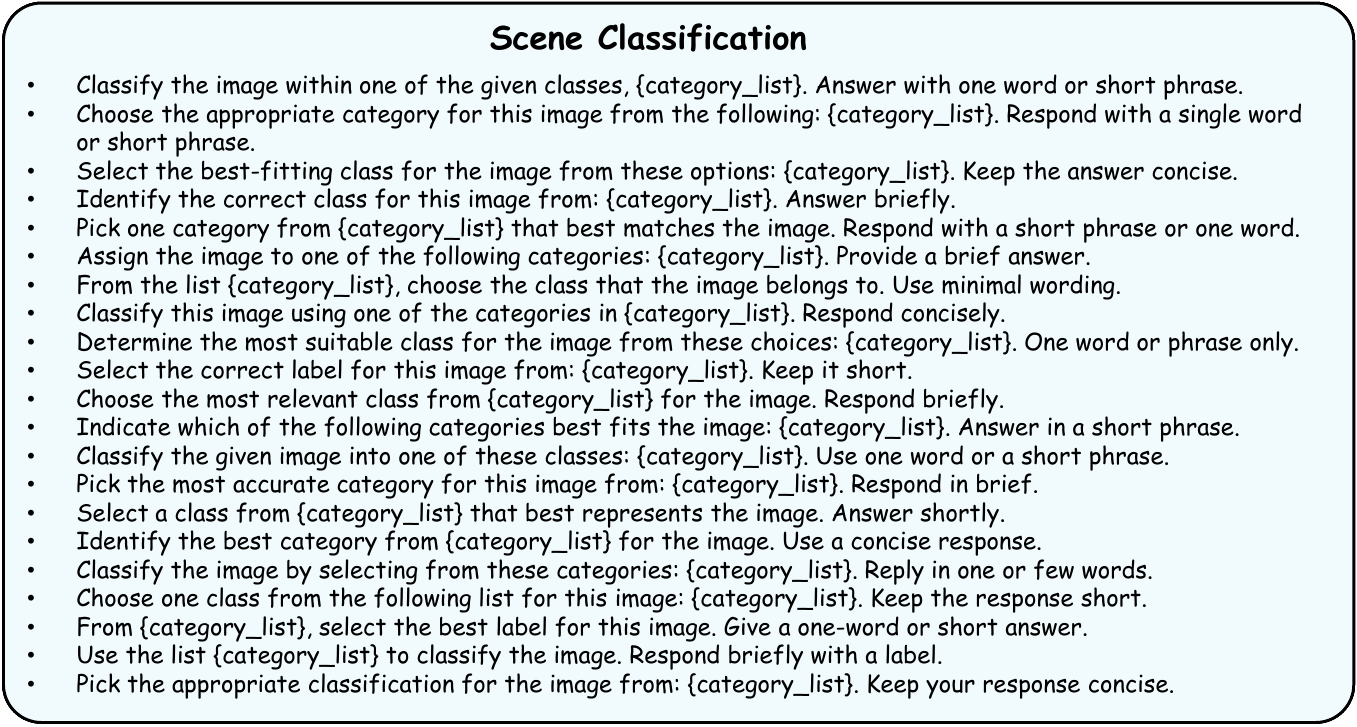}
    \caption{Textual hints for scene classification tasks in the construction of the GeoZero-Raw dataset.}
    \label{fig:hint_sc}
\end{figure*}

\begin{figure*}
    \centering
    \includegraphics[width=\linewidth]{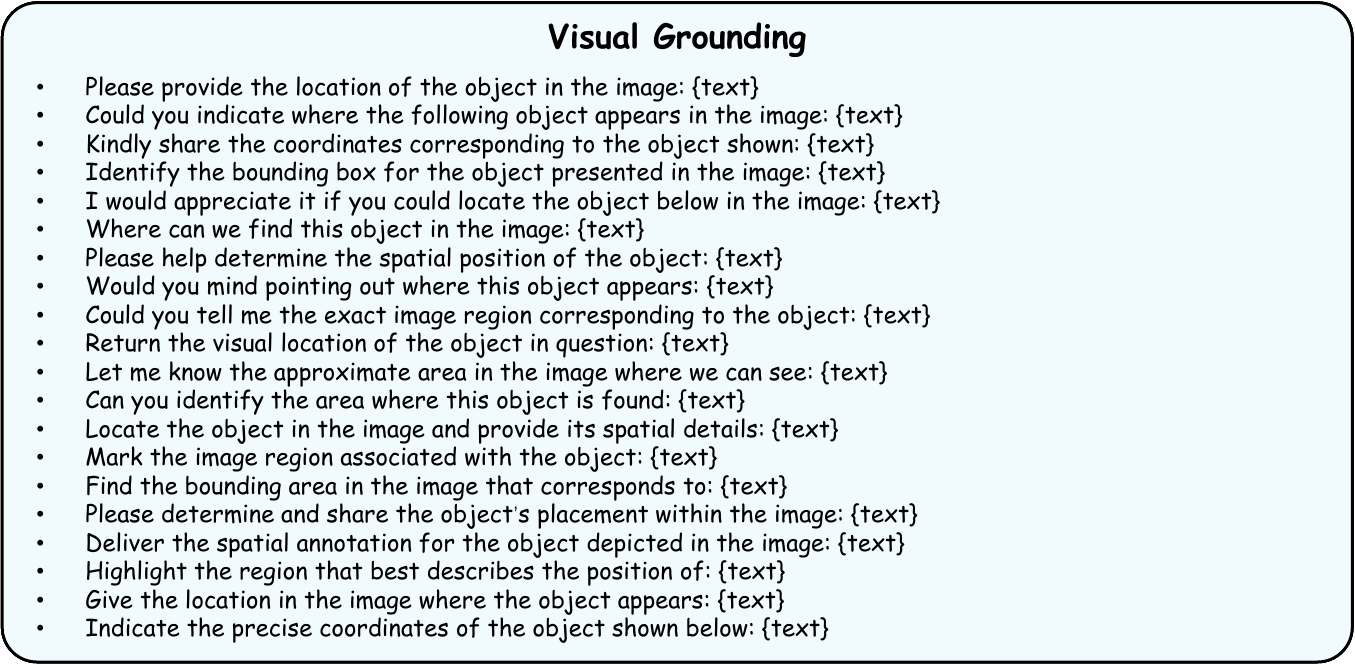}
    \caption{Textual hints for visual grounding tasks in the construction of the GeoZero-Raw dataset.}
    \label{fig:hit_vg}
\end{figure*}

\begin{figure*}
    \centering
    \includegraphics[width=\linewidth]{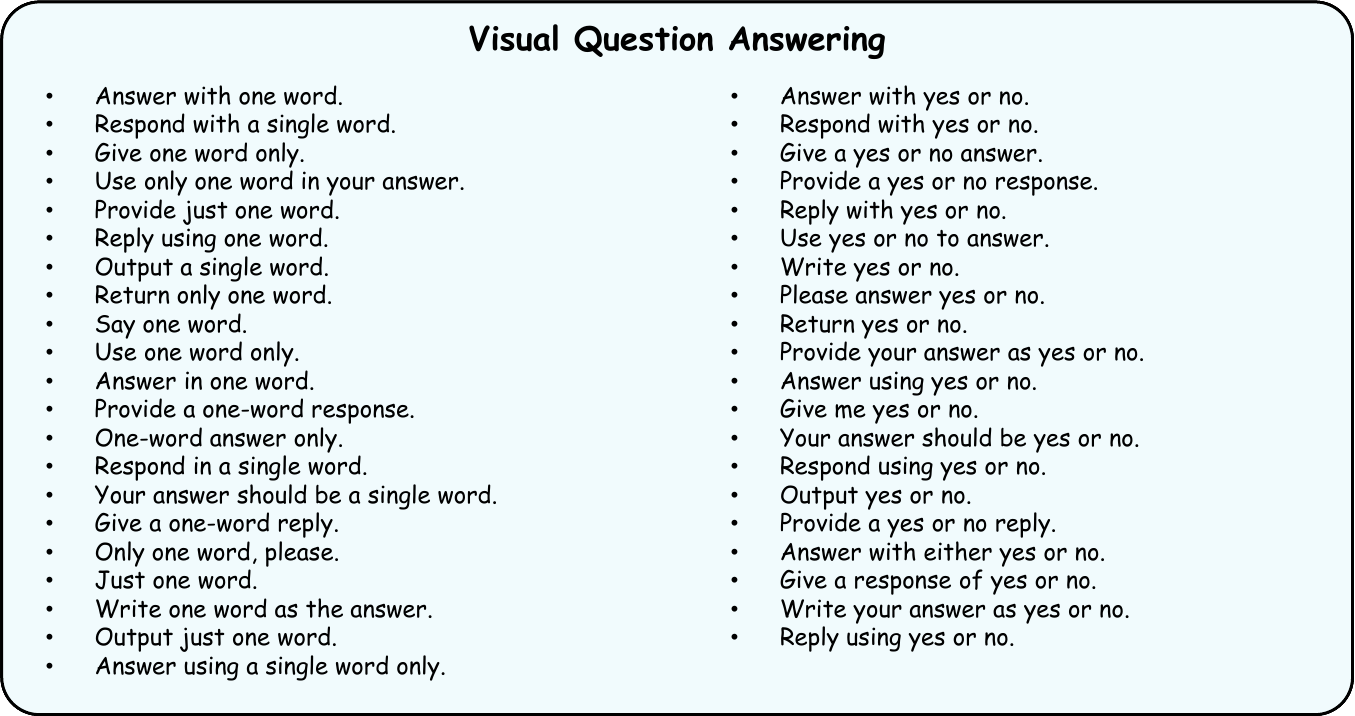}
    \caption{Textual hints for visual question answering tasks in the construction of the GeoZero-Raw dataset.}
    \label{fig:hint_vqa}
\end{figure*}

\begin{figure*}
    \centering
    \includegraphics[width=\linewidth]{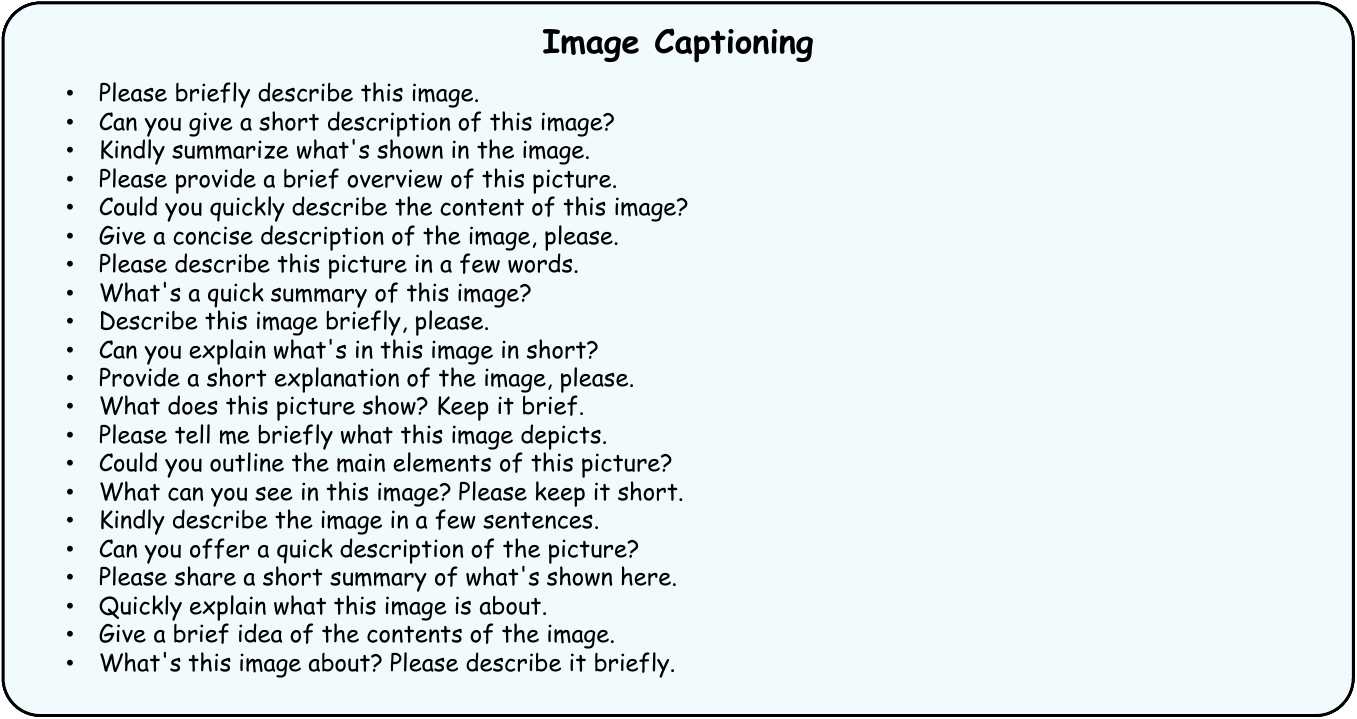}
    \caption{Textual hints for image captioning tasks in the construction of the GeoZero-Raw dataset.}
    \label{fig:hint_ic}
\end{figure*}

\begin{figure*}
    \centering
    \includegraphics[width=0.9\linewidth]{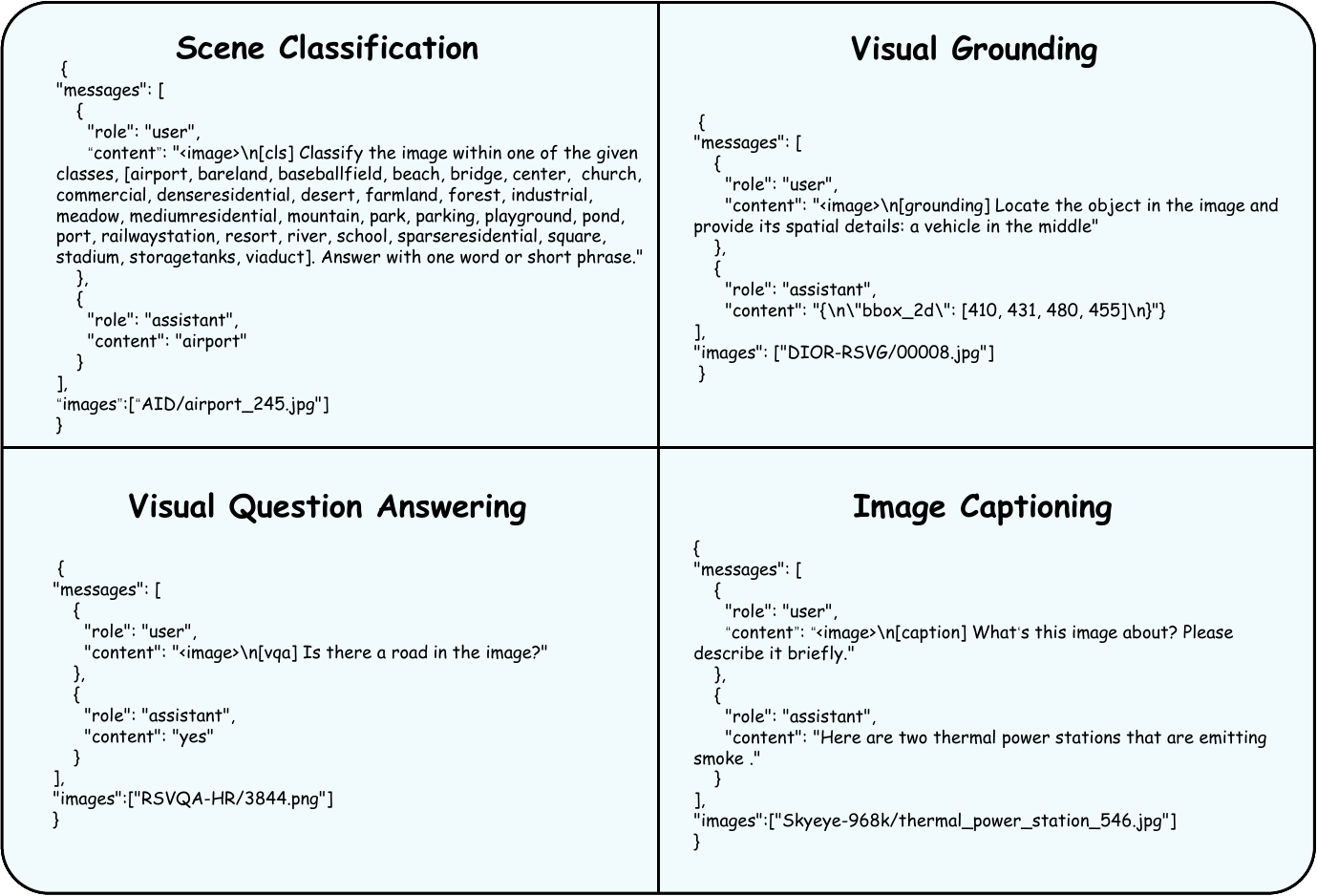}
    \caption{Examples of four task types from the GeoZero-Raw dataset after instruction formatting.}
    \label{fig:train_example}
\end{figure*}

We apply instruction formatting to the GeoZero-Raw dataset. To distinguish different tasks, following the conventions of existing remote sensing MLLMs \cite{kuckreja2024geochat,zhan2025skyeyegpt}, we prepend task descriptors to user instructions, where \textit{[cls]}, \textit{[grounding]}, \textit{[vqa]}, and \textit{[caption]} represent scene classification (SC), visual grounding (VG), visual question answering (VQA), and image captioning (IC) tasks, respectively. For samples that do not belong to these four vision–language tasks, no task descriptors are added. Furthermore, to help general-purpose MLLMs better understand the expected output format for different remote sensing vision–language tasks, we introduce task-specific textual hints in addition to the original user instructions.. For example, for VG tasks, the instruction is reformulated as: ``\textit{Please provide the location of the object in the image: \{original grounding text\}.}'' To prevent the model from over-relying on these textual hints, we randomly decide whether to include them with a probability of 50\%. In addition, to further enhance model generalization, we design more than 20 variations of textual hints for each task, from which one is randomly selected for each sample. The textual hints for different tasks are shown in Figure \ref{fig:hint_sc}-\ref{fig:hint_ic}, while the converted samples are presented in Figure \ref{fig:train_example}.

\section{More Details in Producing GeoZero-Hard}
\label{sec:produce_hard}

\noindent\textbf{Data Splitting.} 
To reduce potential bias and mitigate trivial memorization effects, we first randomly sample approximately 300K instances from GeoZero-Raw to form a validation subset, denoted as \textbf{GeoZero-Raw-val}, which is reserved exclusively for hard-sample selection.
The remaining samples constitute \textbf{GeoZero-Raw-train} and are used solely for training.

This split ensures that the difficulty assessment is performed on samples unseen during subsequent training.

\noindent\textbf{Data Filtering Model.} The data filtering model (DFM) is trained via supervised fine-tuning on the GeoZero-Raw-train dataset, using Qwen2.5-VL-7B-Instruct \cite{qwen25-vl} as the base model. Training is conducted for 3 epochs with a global batch size of~32 across 8~NVIDIA A100 GPUs. We use a learning rate of~1e-4 with a weight decay of~0.1. The vision encoder is kept frozen, while the remaining modules are fine-tuned with LoRA \cite{lora} (rank~8, $\alpha=32$). We adopt bfloat16 precision, FlashAttention~\cite{dao2022flashattention}, and DeepSpeed~\cite{deepspeed} to reduce memory usage and improve training efficiency.

\noindent\textbf{First-Stage Filtering.} 
In the first stage, we apply the trained DFM to all samples in GeoZero-Raw-val that belong to the SC, VG, VQA, and IC tasks, and perform a single forward inference pass. 

Samples that are predicted incorrectly are retained for further consideration. Here, we emphasize that misprediction is not assumed to be a definitive indicator of reasoning complexity; rather, it serves as a practical proxy for identifying samples that are challenging for a strong instruction-tuned MLLM.

The task-specific criteria for judging prediction correctness are defined as follows:

For SC, we treat a prediction as correct if the normalized predicted label exactly matches the normalized ground-truth label, where normalization removes punctuation, lowercases the text, and collapses multiple spaces, yielding a clean, standardized textual form for comparison.

For VG, we evaluate correctness using the Intersection-over-Union (IoU) between the predicted and ground-truth bboxes. A prediction is considered correct if IoU~$\ge$0.5.

For VQA, we adopt a permissive matching criterion: after applying the same normalization, a prediction is deemed correct if it exactly matches the ground-truth answer or if either string is a substring of the other, reducing unnecessary penalties caused by minor phrasing discrepancies.

For IC, we compute a word-level $F_1$ score between the predicted and reference captions. After normalization, both captions are split into words and converted into unique word sets. Precision and recall are then computed based on set overlap, and the $F_1$ score is calculated as their harmonic mean. A caption is considered correct if $F_1$ $\ge$ 0.6.

All samples that are predicted incorrectly across these four tasks are treated as potential hard examples and carried into the second-stage filtering.

\noindent\textbf{Second-Stage Filtering.} For each candidate hard sample, we perform three stochastic generations using the DFM and evaluate each output with the corresponding task-specific criterion, yielding a binary correctness score. The per-sample accuracy $Acc$ is the average correctness across the three trials, and the difficulty score is defined as $1 - Acc$.

Then, for each task, we rank all samples by their difficulty scores and select the hardest subsets while maintaining balance across tasks. We also perform deduplication to remove redundant samples before finalizing GeoZero-Hard.

\section{Prompt Template for GeoZero-Hard}
\label{sec:prompt_template}

\begin{figure*}
    \centering
    \includegraphics[width=0.87\linewidth]{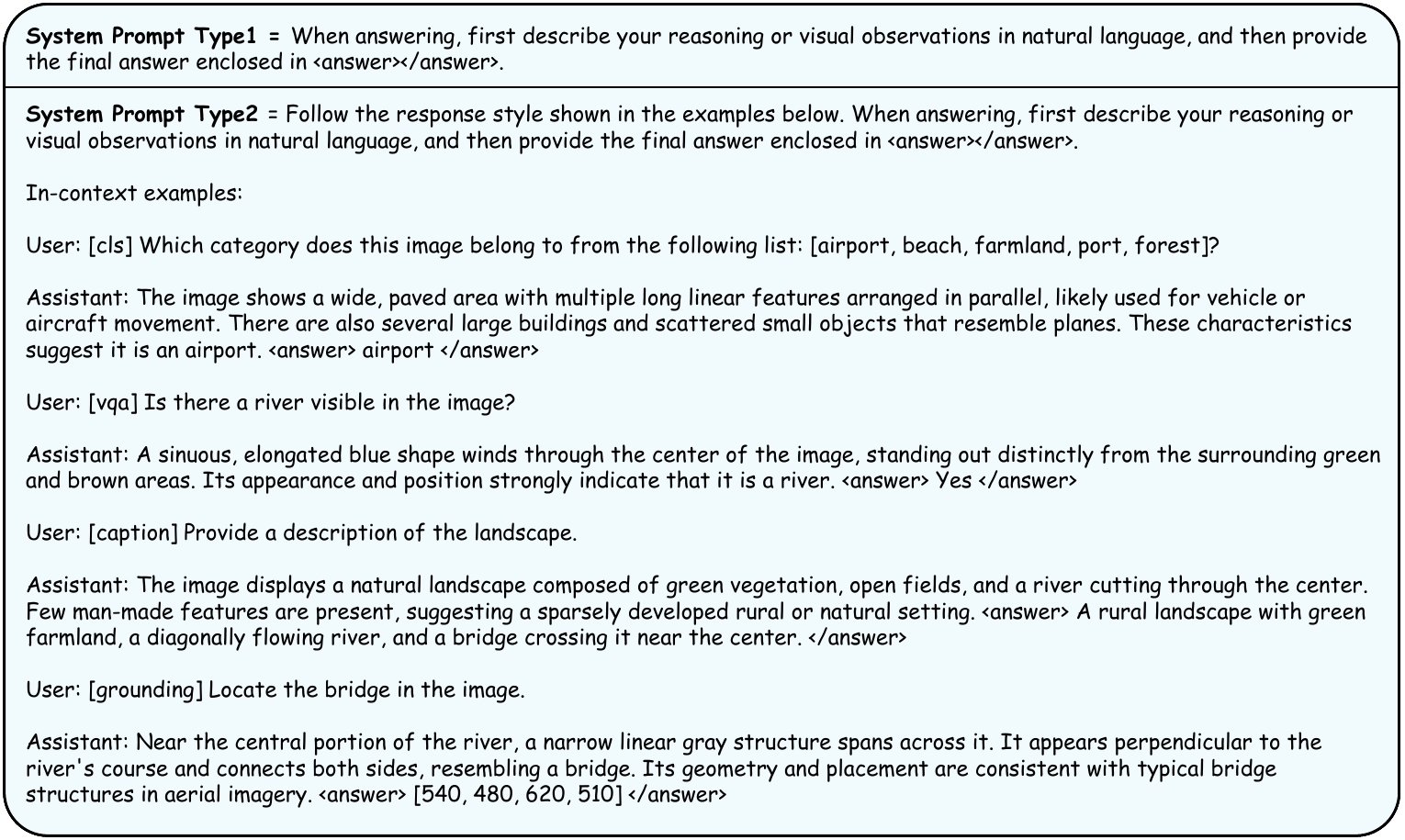}
    \caption{System prompts in the GeoZero-Hard dataset. Type1: without hypothetical examples. Type2: with hypothetical examples.}
    \label{fig:system_prompt}
\end{figure*}

During the RL stage, we introduce system prompts to the GeoZero-Hard dataset. To help the model understand the expected answer style for each task, we incorporate a single hypothetical answer example into the system prompt. Similar to the textual hints used in GeoZero-Raw, we randomly decide whether to include these illustrative examples with a probability of 50\% to enhance model generalization. We argue that such examples do not introduce human bias, as they are entirely hypothetical and no corresponding images are provided, they merely serve to clarify the expected output format for the model. The specific examples of these system prompts are presented in Figure \ref{fig:system_prompt}.

\section{Thinking-Quality Score $s_t$}
\label{sec:think_quality_score}

Formally, for the thinking-quality score $s_t$, we define:
\begin{equation}
s_t = (1 - w_d) \cdot q_t + w_d \cdot (q_t \cdot b_d),
\label{eq:thinking_score}
\end{equation}
where \( q_t \) represents the base structural score, capturing the intrinsic reasoning quality by integrating the effects of trajectory length, redundancy, and answer overlap.  
\( b_d \) denotes the semantic diversity bonus of the reasoning text. The coefficient \( w_d \) controls the contribution of semantic diversity to the overall thinking score.  In our implementation, $w_d$ is set to 0.3. It is worth noting that semantic diversity is meaningful only when the structural quality of reasoning is sufficiently high. Therefore, \( b_d \) is multiplied by \( q_t \) to adaptively modulate its influence.

Furthermore, the structural score \( q_t \) is defined as:
\begin{equation}
q_t = l_s \cdot p_r \cdot p_a,
\label{eq:structure_score}
\end{equation}
where \( l_s \in [0,1] \) denotes the normalized length score of the reasoning trajectory, encouraging adequate but not excessively long reasoning texts;
\( p_r \in [0,1] \) is the redundancy penalty that suppresses repetitive or verbose expressions;  
and \( p_a \in [0,1] \) represents the answer-overlap penalty, which discourages the model from directly copying the final answer into the reasoning text.  

Equations~(\ref{eq:thinking_score})-(\ref{eq:structure_score}) jointly ensure that only reasoning processes that are structurally sound, sufficiently informative, semantically diverse, and clearly distinct from the final answer can achieve high thinking-quality scores. The specific computation of $l_s$, $p_r$, $p_a$ and $b_d$ are provided as follows:

\noindent\textbf{Normalized Length Score.}
Let $T$ denote the reasoning text and $w(T)$ its word count.
The normalized length score $l_s$ is computed according to the following formulation:
\begin{equation}
l_s(T) =
\begin{cases}
0 & w(T) < \tau_{\min},\\[3pt]
\dfrac{w(T) - \tau_{\min}}{\tau_{\mathrm{lo}} - \tau_{\min}} & \tau_{\min} \le w(T) < \tau_{\mathrm{lo}},\\[8pt]
1 & \tau_{\mathrm{lo}} \le w(T) \le \tau_{\mathrm{hi}},\\[3pt]
1 - \dfrac{w(T) - \tau_{\mathrm{hi}}}{\tau_{\max} - \tau_{\mathrm{hi}}} & \tau_{\mathrm{hi}} < w(T) \le \tau_{\max},\\[8pt]
0 & w(T) > \tau_{\max}.
\end{cases}
\label{eq:length_score}
\end{equation}

\noindent
Here, $\tau_{\min}$ specifies the minimum acceptable reasoning length,
$\tau_{\mathrm{lo}}$ marks the point at which the score first reaches saturation,
$\tau_{\mathrm{hi}}$ denotes the upper boundary of the plateau region,
and $\tau_{\max}$ defines the maximum tolerable length beyond which the score decays back to zero. This design encourages sufficiently detailed reasoning while suppressing trajectories that are either too short or excessively long. In our implementation, $\tau_{\min}$, $\tau_{\mathrm{lo}}$, $\tau_{\mathrm{hi}}$, and $\tau_{\max}$ are set to 20, 40, 80, and 160, respectively.

\noindent\textbf{Redundancy Penalty.}
To measure redundancy, we first define the redundancy ratio of $T$ as
\begin{equation}
d(T) =
\frac{\bigl|\{\,u \in V_T \mid c(u,T) \ge 2\,\}\bigr|}
{|V_T|},
\label{eq:redund_ratio}
\end{equation}
where $V_T$ denotes the set of distinct words in $T$, and $c(u,T)$ is the occurrence count of word $u$ in $T$.
In addition to word-level repetition, we further detect generic filler expressions using a predefined phrase set $\mathcal{Z}$. The redundancy penalty $p_r$ is then defined as
\begin{equation}
p_r(T) =
\begin{cases}
\gamma_r &
d(T) > \delta_r \ \text{or}\
\exists\, z \in \mathcal{Z}\ \text{s.t.}\ z \in T,\\[3pt]
1 & \text{otherwise},
\end{cases}
\label{eq:redund_penalty}
\end{equation}
where $\delta_r$ is the redundancy threshold (set to 0.15) and $\gamma_r \in (0,1)$ is a down-weighting factor (set to $\gamma_r$ = 0.5 in our implementation).
The phrase set $\mathcal{Z}$ contains common templated expressions, including
\textit{\{"therefore", "however", "in conclusion", "overall", "in summary"}\},
which typically indicate overly formulaic reasoning.

\noindent\textbf{Answer-Overlap Penalty.}
Similarly, to discourage copying the final answer into the reasoning text, we compute an answer-overlap ratio.
Let $A$ denote the predicted answer text.  
The overlap ratio is defined as:
\begin{equation}
o(T, A) =
\frac{\bigl|\{\,u \in A \mid u \in T\,\}\bigr|}{|A|}.
\end{equation}
The answer-overlap penalty $p_a$ is then given by
\begin{equation}
p_a(T, A) =
\begin{cases}
1 & o(T, A) \le \tau_a,\\[6pt]
1 - \dfrac{o(T, A) - \tau_a}{1 - \tau_a} & o(T, A) > \tau_a,
\end{cases}
\label{eq:ans_overlap_penalty}
\end{equation}
where $\tau_a$ is the overlap gate.  
The penalty activates only when the overlap exceeds $\tau_a$, and its magnitude increases proportionally with the degree of excess.
In our implementation, we set $\tau_a = 0.3$.

\noindent\textbf{Semantic Diversity Bonus.}
For the semantic diversity bonus $b_d$, we first split $T$ into $n$ sentences $\{t_i\}_{i=1}^n$ and obtain their normalized embeddings $\{\mathbf{e}_i\}_{i=1}^n$ using the bge-small-en-v1.5 model~\cite{bge_embedding}.  
We then quantify sentence-level semantic diversity as
\begin{equation}
\rho = 1 - \frac{1}{n-1} \sum_{i=1}^{n-1}
\cos\bigl(\mathbf{e}_i,\,\mathbf{e}_{i+1}\bigr),
\end{equation}
which increases when neighboring sentences exhibit lower semantic similarity.  
Finally, we have
\begin{equation}
b_d = \text{clip}(\rho,\,0,\,1),
\label{eq:diversity_bonus}
\end{equation}
which maps the score to $[0,1]$.  The clipping step ensures that the bonus reflects meaningful semantic diversity and prevents degenerate or contradictory sentences, whose embeddings may yield artificially low cosine similarities, from receiving disproportionately high reward.

It should be noted that we initially included the answer-overlap penalty out of completeness, as a safeguard against potential reward hacking where the model might insert the final answer into the reasoning text to artificially inflate the thinking reward.  
However, in practice, we found that the remaining components of our reward design, namely the redundancy penalty, the diversity bonus, the length score, and the answer-gated thinking reward, already suppress such behavior reliably.  
As a result, this term plays only a marginal role during training.

Furthermore, since correct reasoning naturally incorporates answer-related words, the overlap ratio tends to be high across most samples.  
This produces an almost uniform scaling effect that does not alter the relative reward differences between candidate trajectories.  
Therefore, both the training dynamics and the conclusions reported in the main paper remain unaffected.

Interestingly, this may also partly explain why our method achieves especially strong improvements over comparison methods on grounding tasks.  
While gains are observed across all tasks, grounding answers consist of numeric coordinates with essentially no lexical overlap with the reasoning text. As a result, this near-uniform penalization has limited influence on the thinking trajectories in VG tasks.

\section{Task-Specific Answer Reward $r_a$}

\label{sec:task_answer_reward}

For each task, we define a task-specific answer reward $r_a$ that lies in $[0,1]$.

\noindent\textbf{Scene Classification and Visual Question Answering.}
For scene classification (SC) and visual question answering (VQA), we employ a text embedding model $f$ to obtain feature representations of the predicted answer $x$ and the ground-truth answer $y$, and measure their cosine similarity.
The similarity is then mapped from $[-1,1]$ to $[0,1]$ by an affine transformation:
\begin{equation}
    r_a^{(\text{SC})}
    =
    r_a^{(\text{VQA})}
    =
    \frac{\cos\bigl(f(x), f(y)\bigr) + 1}{2},
    \label{eq:sc_vqa}
\end{equation}
where
\(
\cos(f(x), f(y))
=
\frac{f(x) \cdot f(y)}{\lVert f(x) \rVert \, \lVert f(y) \rVert}.
\)
In our implementation, $f$ is instantiated by the bge-small-en-v1.5 model~\cite{bge_embedding} with normalized embeddings.

\noindent\textbf{Visual Grounding.}
For visual grounding (VG), we directly use the Intersection-over-Union (IoU) between the predicted bounding box $\hat{B}$ and the ground-truth box $B$ as the answer reward:
\begin{equation}
    r_a^{(\text{VG})} = \mathrm{IoU}(\hat{B}, B).
\end{equation}

\noindent\textbf{Image Captioning.}
For image captioning (IC), we combine multiple complementary metrics to provide a comprehensive assessment of caption quality.  
Given a predicted caption and a reference caption, we compute four components:

\begin{itemize}

\item[\textbf{(1)}] \textbf{BLEU score.}
We compute BLEU-1 to BLEU-4 using the standard $n$-gram precision formulation with smoothing, 
and aggregate them via a weighted geometric mean:
\begin{equation}
\begin{aligned}
s_b = \exp\Bigl(
    &0.1 \log(\text{BLEU-1})
   + 0.2 \log(\text{BLEU-2})
   \\
   &+ 0.3 \log(\text{BLEU-3})
   + 0.4 \log(\text{BLEU-4})
\Bigr).
\end{aligned}
\end{equation}
This emphasizes higher-order $n$-gram matches and captures word-level overlap and local fluency.

\item[\textbf{(2)}] \textbf{ROUGE score.}
We compute ROUGE-1 and ROUGE-L F-scores, and the final ROUGE score is obtained by averaging them:
\begin{equation}
s_r = \frac{1}{2}\left(\text{ROUGE-1} + \text{ROUGE-L}\right).
\end{equation}
This captures lexical overlap and global structural consistency between the predicted and reference captions.

\item[\textbf{(3)}] \textbf{METEOR score.}  
We directly use the METEOR score between the predicted and reference captions as
\begin{equation}
s_m = \text{METEOR}.
\end{equation}
This metric is sensitive to synonymy, stemming, and paraphrasing, and thus provides a more semantically aligned measure of caption quality.

\item[\textbf{(4)}] \textbf{Semantic similarity score.}  
We additionally include an embedding-based semantic similarity score $s_e$ that evaluates the semantic alignment between the predicted and reference captions.  
This score is computed in the same way as the embedding-based answer reward used for SC and VQA tasks (see Eq.~(\ref{eq:sc_vqa})).

\end{itemize}

The final caption reward is obtained via a weighted sum:
\begin{equation}
    r_a^{(\text{IC})}
    =
    w_b s_b
    + w_r s_r
    + w_m s_m
    + w_e s_e,
\end{equation}
where $w_b, w_r, w_m, w_e$ control the contribution of the four scores and satisfy  
$w_b + w_r + w_m + w_e = 1$.  
In our implementation, we set $w_b$, $w_r$, $w_m$ and $w_e$ to
0.15, 0.2, 0.3, and 0.35, respectively.

\begin{table}[t]
\scriptsize
\caption{Training hyperparameters for evaluating on different datasets used in scene classification, visual grounding, visual question answering, image captioning tasks and comprehensive benchmarks with GeoZero (w/ RFT).}
\centering
\resizebox{\linewidth}{!}{
\begin{tabular}{lccc}
\toprule
Dataset & Epoch & Learning Rate & Global Batch Size \\
\midrule 
UCM \cite{ucm}           &  1 & 5e-5 & 144 \\
RSVG \cite{rsvg}             &  1 & 5e-5 & 48 \\
\bottomrule
\end{tabular}}
\label{tab:train_hyperparams}
\end{table}

\section{Reinforcement Fine-tuning Settings}

\label{sec:rft_settings}

The hyperparameters used for reinforcement fine-tuning on different datasets are summarized in Table~\ref{tab:train_hyperparams}. All other settings follow the reinforcement learning stage described in the main paper.

\section{Further Investigation of Reasoning Behavior}
\label{sec:analysis_reason}

\subsection{Overall Characteristics of Model Reasoning.} 

Figure~\ref{fig:think_vs_answer_supp}(a)–(f) illustrates the distribution of average sample accuracy across different reasoning-related metrics. It should be noted that in the main paper, we conducted a preliminary analysis using two variables: the word-level reasoning length and the thinking-quality score~$s_t$ (see (a) and (f)). Here, we take a closer look by further decomposing $s_t$ into its internal factors. Specifically, we additionally analyze four variables: the normalized length score~$l_s(T)$, the non-redundancy ratio~$1-d(T)$ (which serves as a redundancy penalty in $q_t$), the structural thinking score~$q_t$, and the semantic diversity bonus~$b_d$. Furthermore, we examine how the sample frequency is distributed across different value intervals of each metric (see (g)–(l)), which helps reveal the overall distribution of reasoning characteristics within the test data. Next, we provide a detailed analysis of each metric individually.

\begin{itemize}
    \item[\textbf{(1)}] \textbf{Reasoning Length.}  Figure~\ref{fig:think_vs_answer_supp}(b) shows a positive correlation between $l_s(T)$ and accuracy: samples with more appropriate reasoning lengths tend to yield more reliable predictions, consistent with the findings in the main paper.
Encouragingly, we find that for the vast majority of samples, the predicted reasoning lengths fall within the expected range, as shown in Figure~\ref{fig:think_vs_answer_supp}(h), where more than 80\% of the samples achieve very high $l_s(T)$ values (greater than~0.8). Figure~\ref{fig:think_vs_answer_supp}(g) further reveals that most reasoning chains have a length between 40 and~80 words, which precisely corresponds to the full-score interval defined in Equation~(\ref{eq:length_score}).

    \item[\textbf{(2)}] \textbf{Non-Redundancy.} As shown in Figures~\ref{fig:think_vs_answer_supp}(c) and (i), almost no accuracy bars appear when $1-d(T) \leq 0.5$, while the vast majority of samples have values above~0.85. This suggests that the model’s reasoning chains exhibit strong non-redundancy across most samples. However, Figure~\ref{fig:think_vs_answer_supp}(c) also indicates that variations in reasoning redundancy have negligible influence on model performance.

\begin{figure*}[t]
    \centering
    \subfigure[]{
        \includegraphics[width=0.3\linewidth]{Figures/dior-rsvg-length_words_binned_mean_acc-v1.png}
        
    }
    \subfigure[]{
        \includegraphics[width=0.3\linewidth]{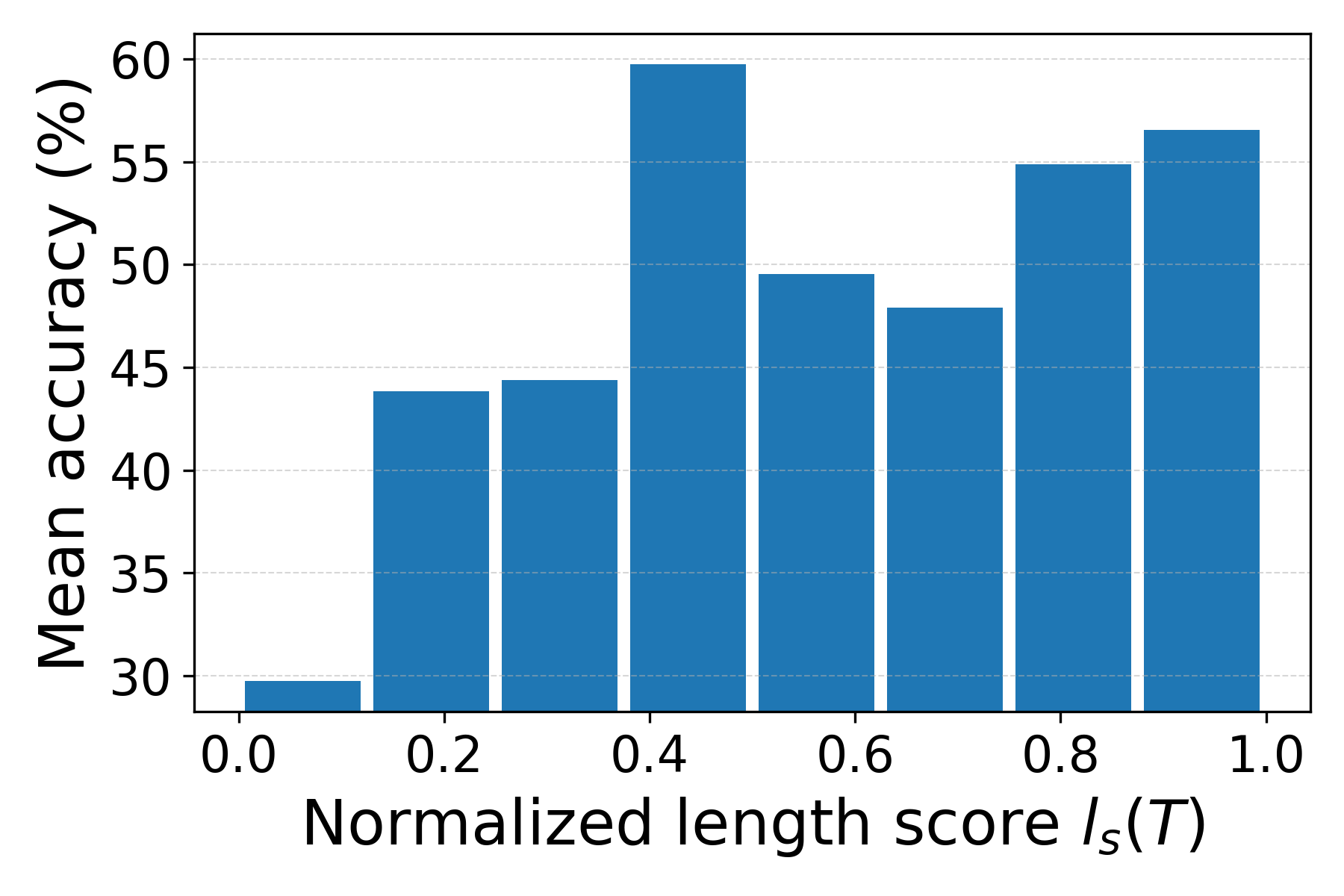}
        
    }
    \subfigure[]{
        \includegraphics[width=0.3\linewidth]{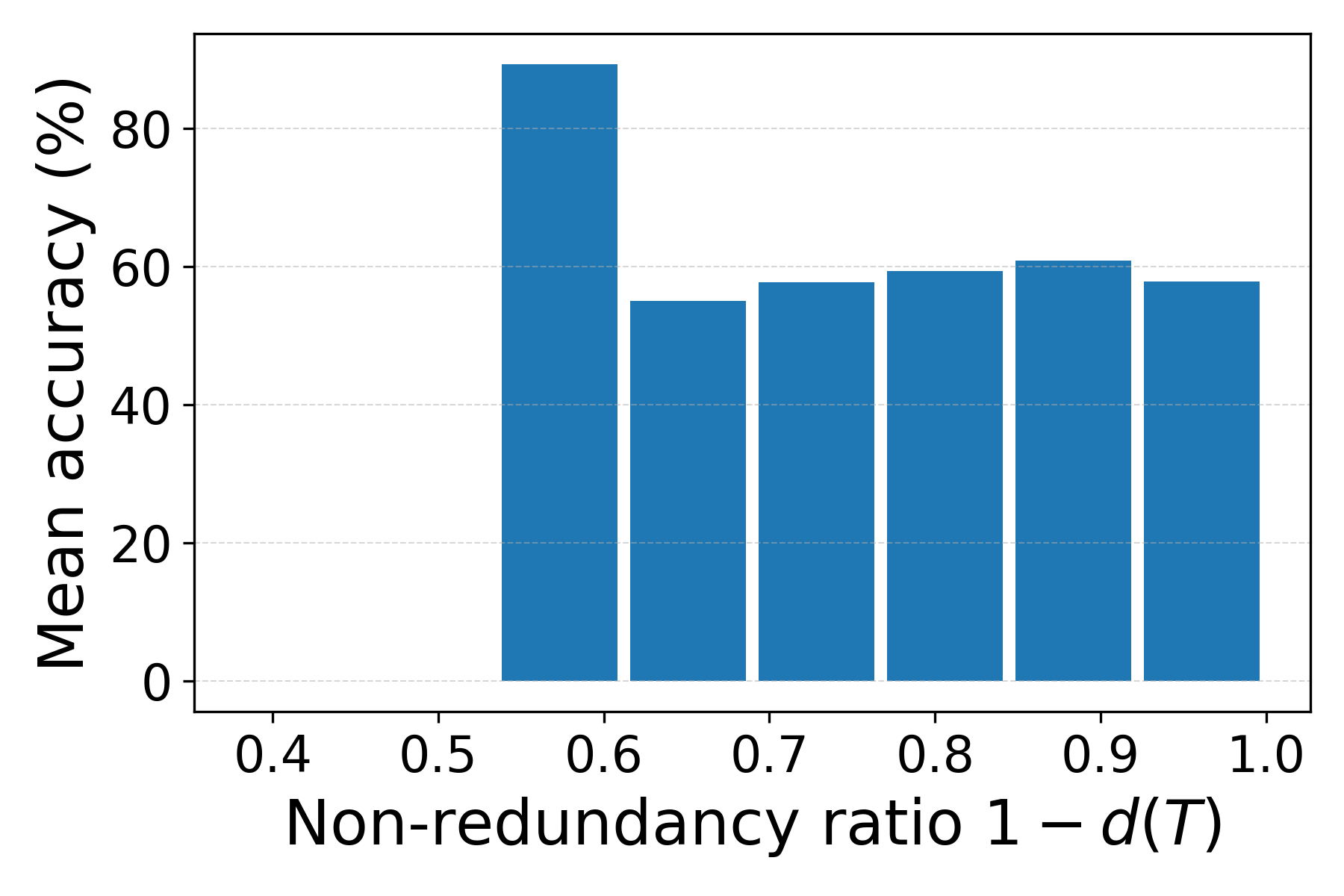}
        
    }
    \\
    \subfigure[]{
        \includegraphics[width=0.3\linewidth]{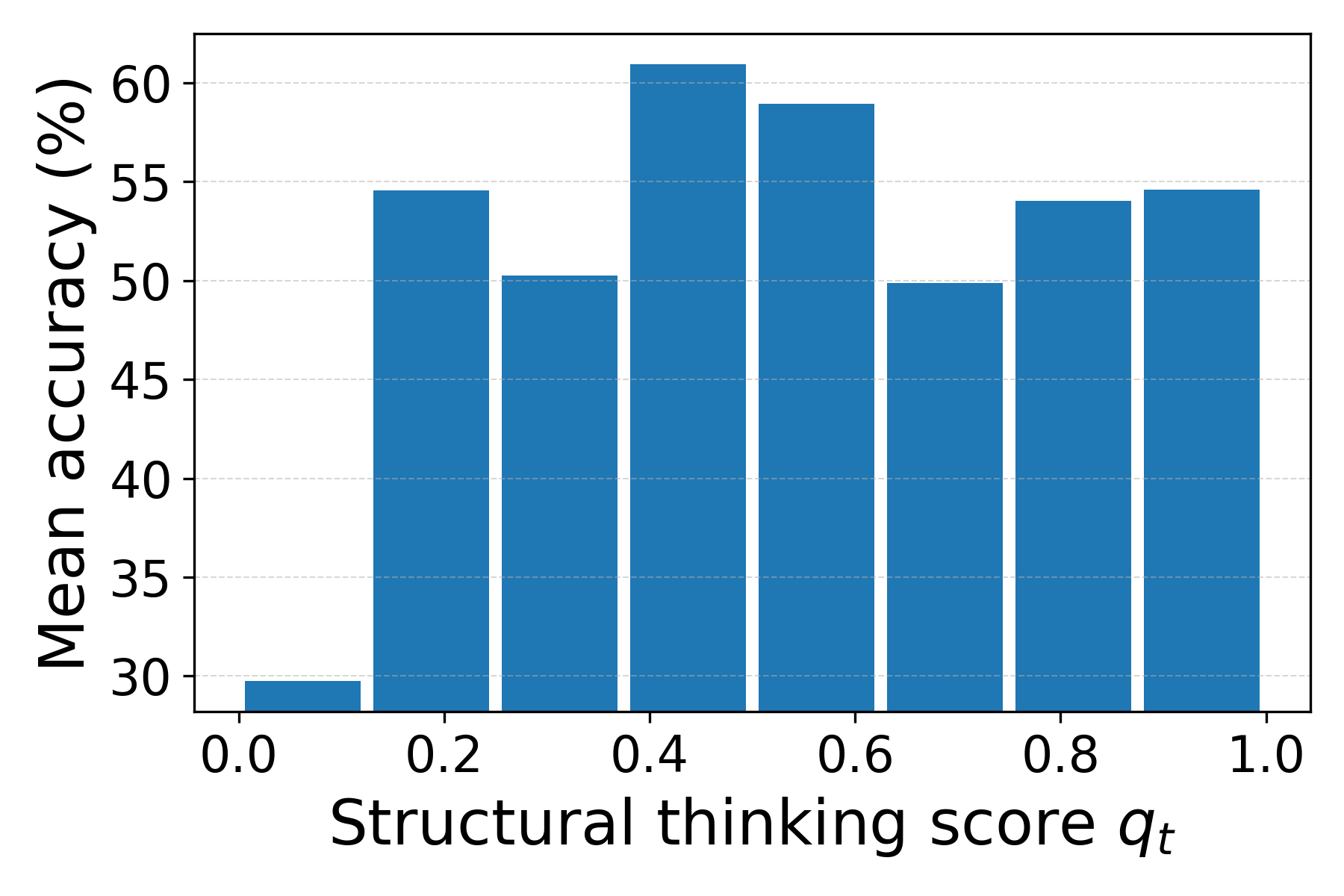}
        
    }
    \subfigure[]{
        \includegraphics[width=0.3\linewidth]{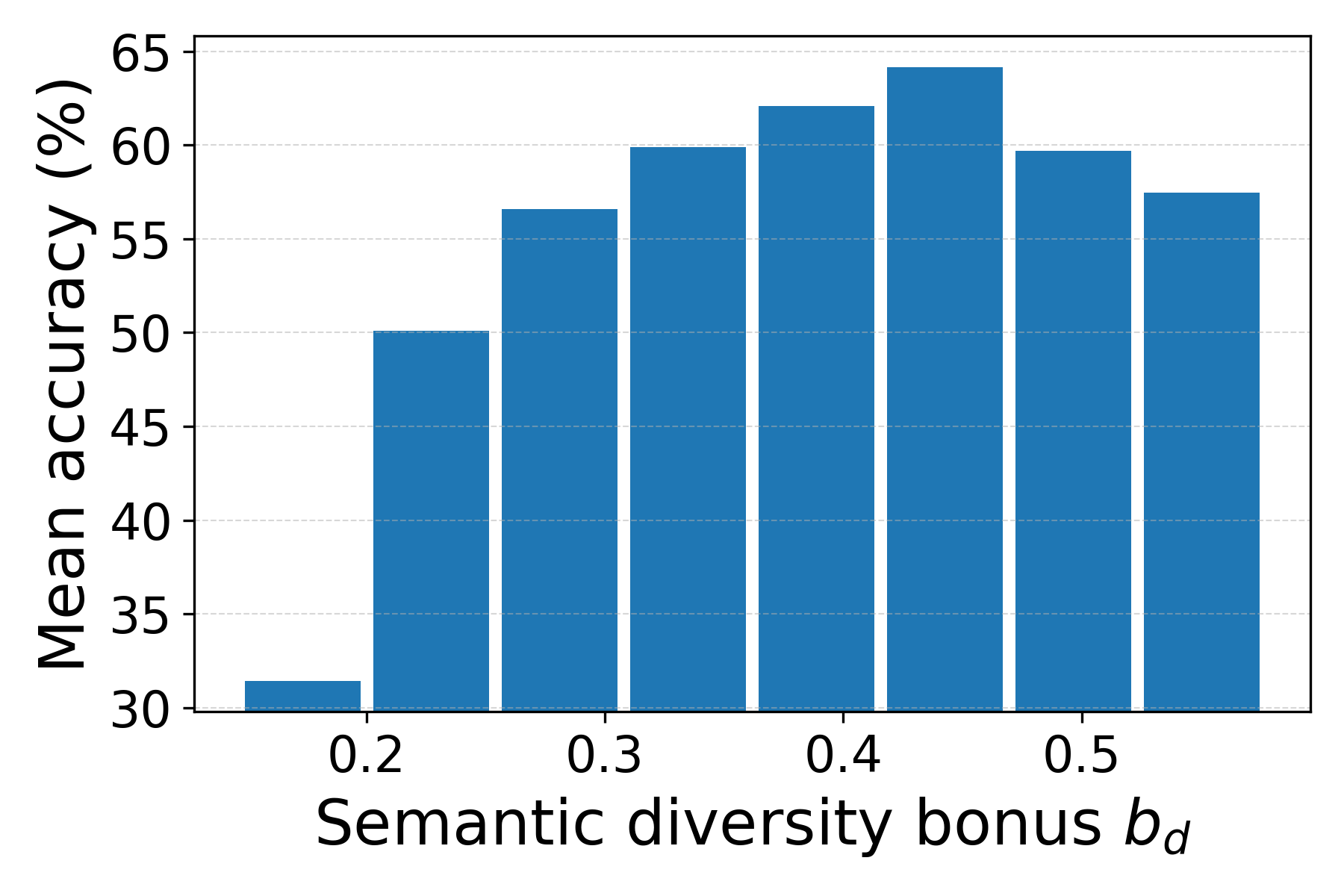}
        
    }
    \subfigure[]{
        \includegraphics[width=0.3\linewidth]{Figures/dior-rsvg-think_bonus_binned_mean_acc-v1.png}
        
    }
    \\
    \subfigure[]{
        \includegraphics[width=0.3\linewidth]{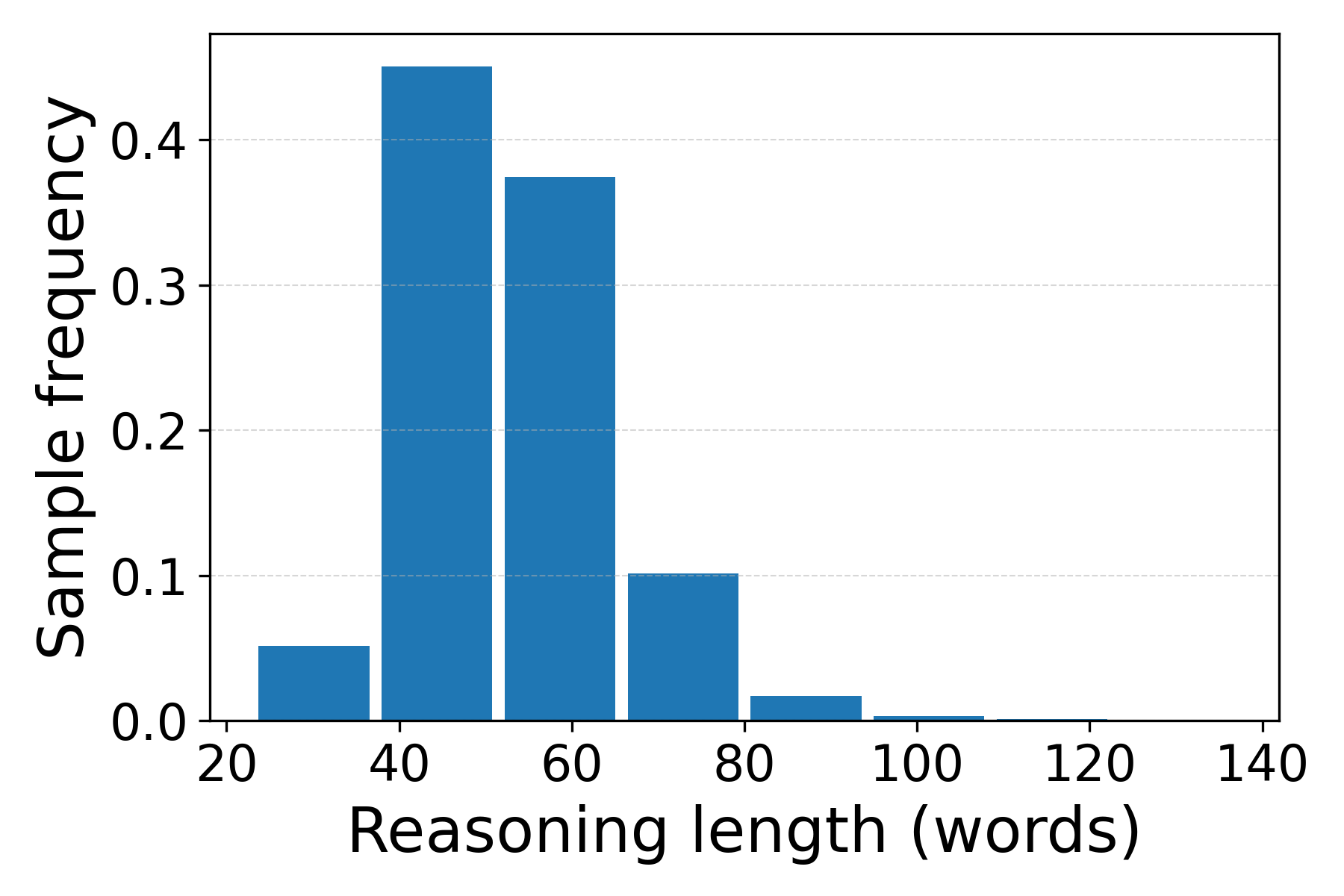}
        
    }
    \subfigure[]{
        \includegraphics[width=0.3\linewidth]{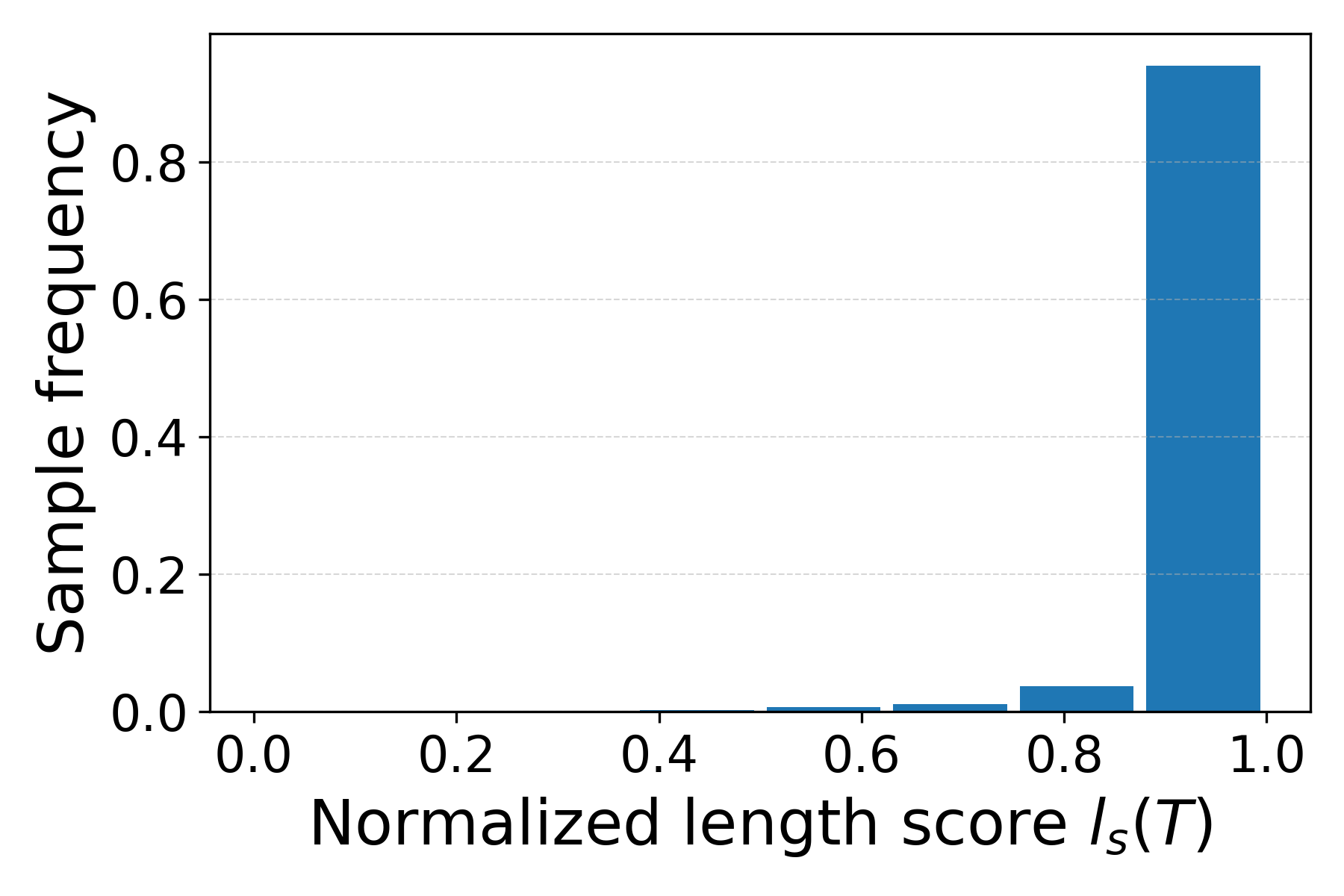}
        
    }
    \subfigure[]{
        \includegraphics[width=0.3\linewidth]{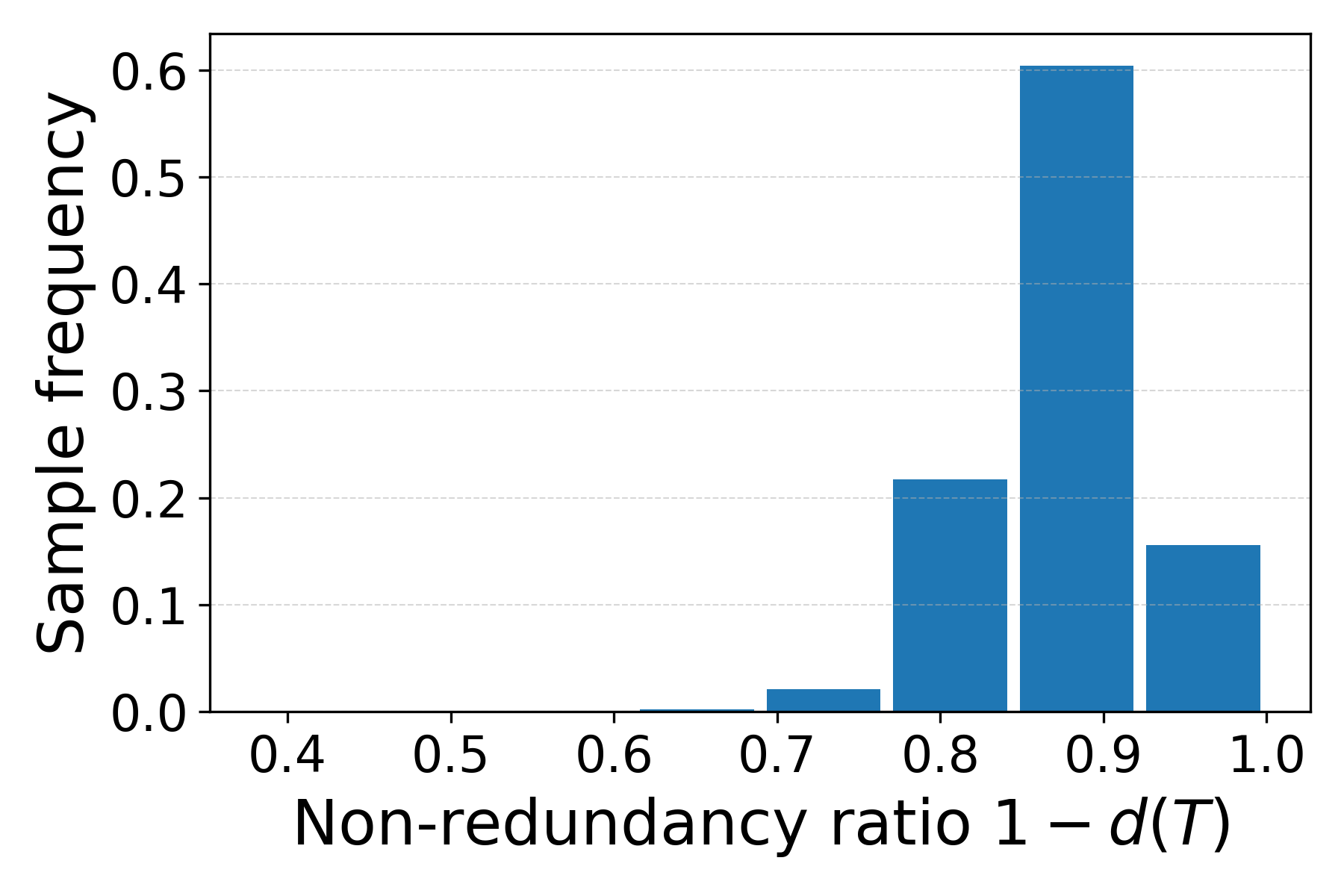}
        
    }
    \\
    \subfigure[]{
        \includegraphics[width=0.3\linewidth]{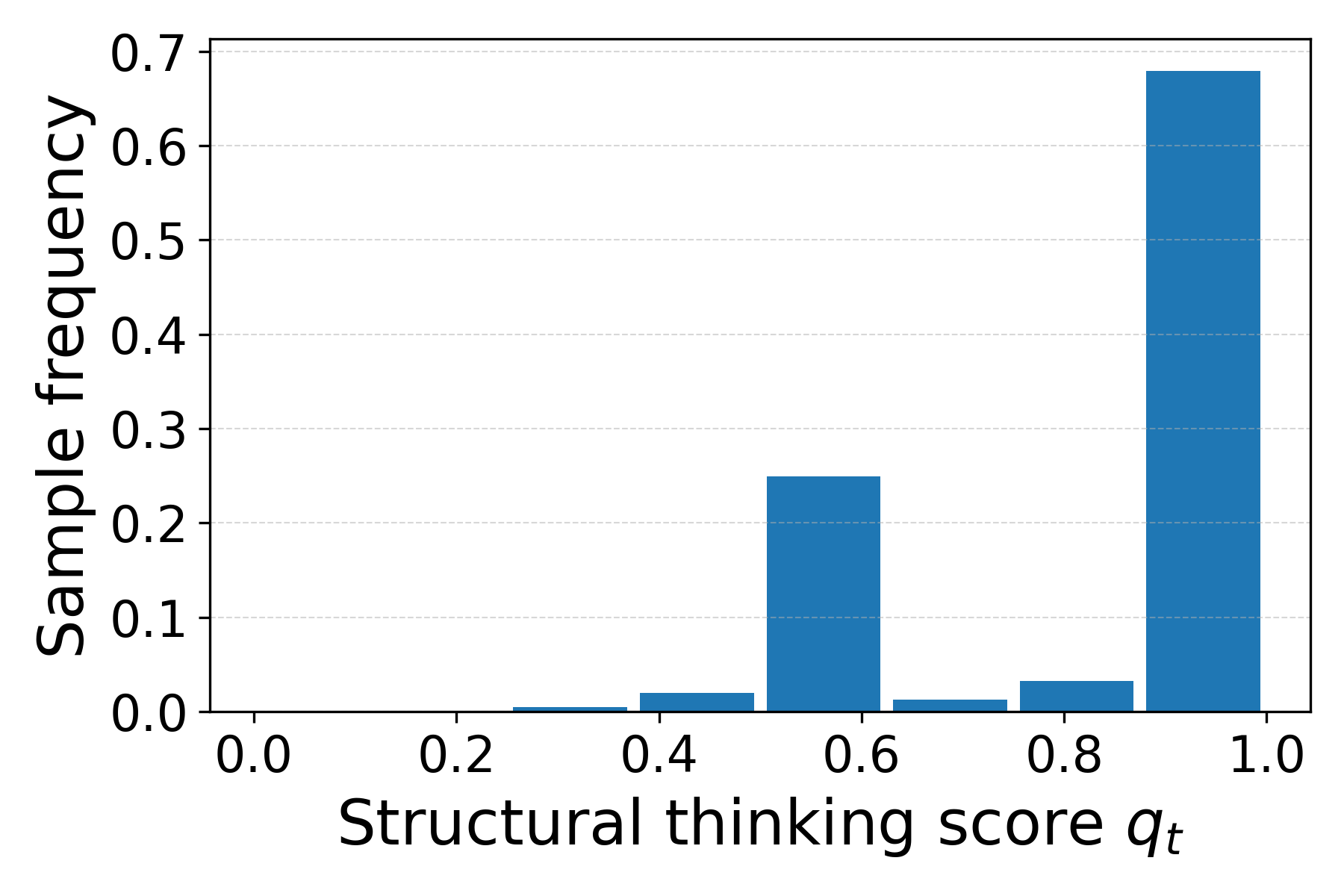}
        
    }
    \subfigure[]{
        \includegraphics[width=0.3\linewidth]{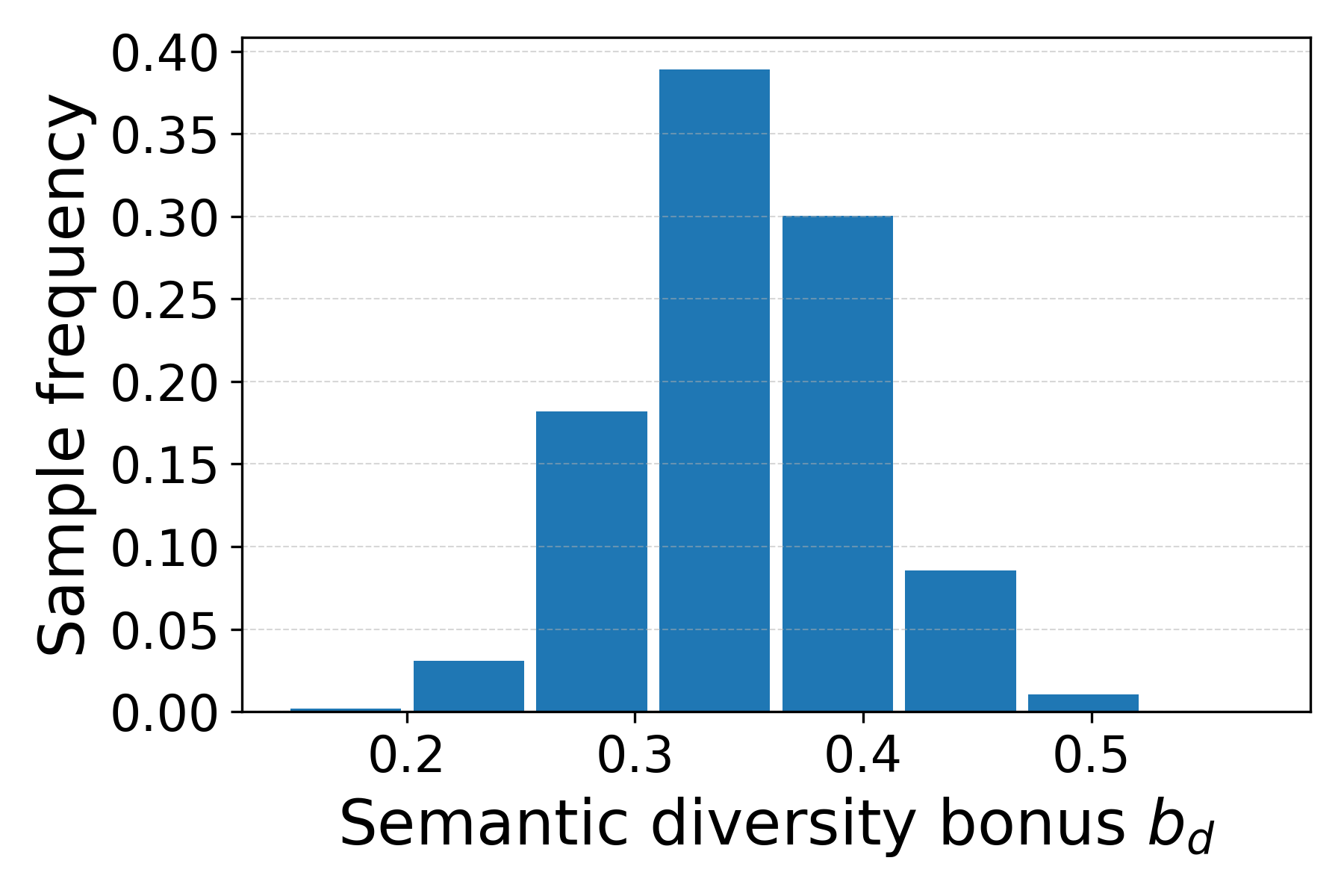}
        
    }
    \subfigure[]{
        \includegraphics[width=0.3\linewidth]{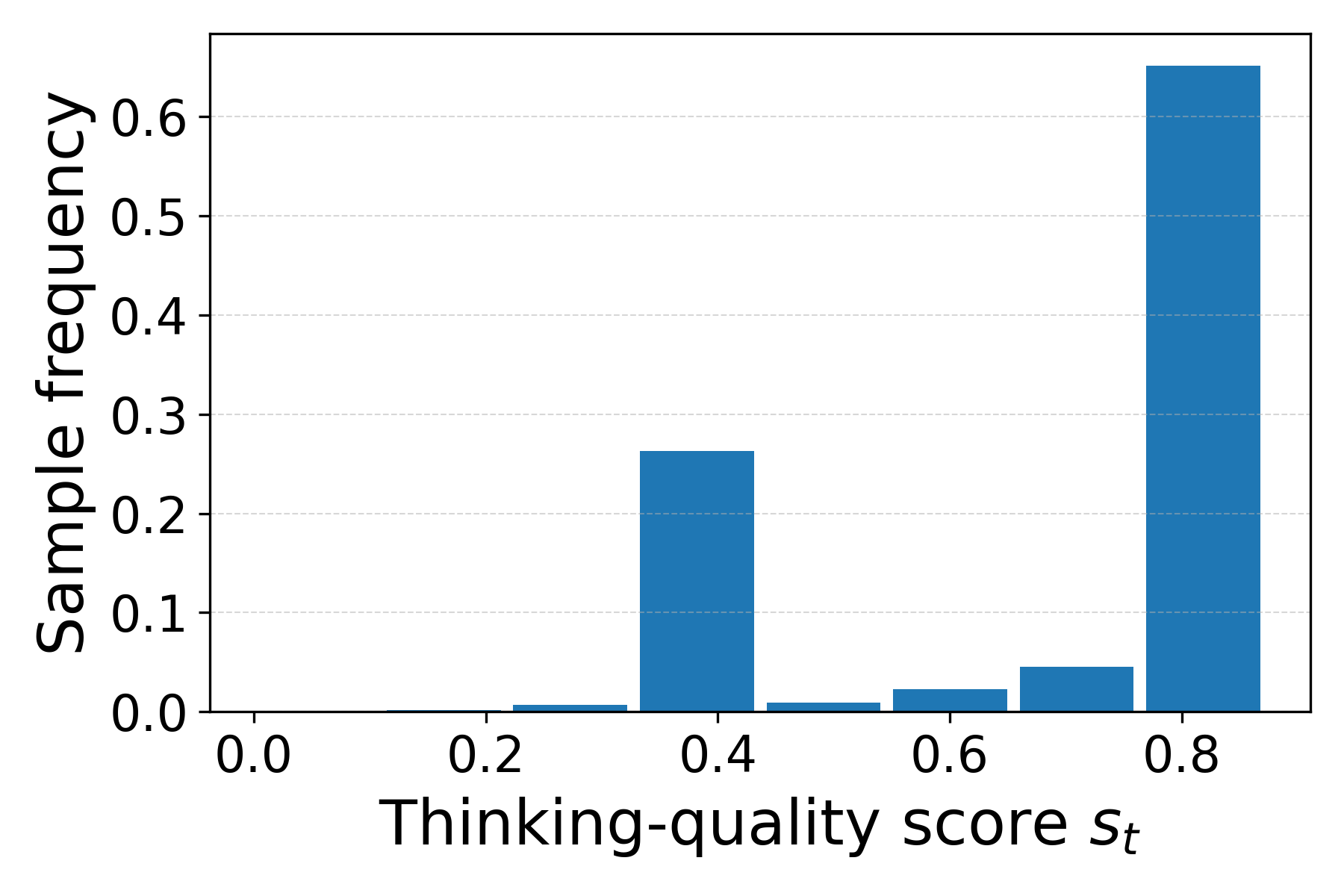}
        
    }
    \caption{The distribution of average accuracy and sample frequency on thinking-related metrics, where the statistics are computed from GeoZero’s predictions on the 7500 samples from the DIOR-RSVG \cite{dior-rsvg} test set. The accuracy of samples are represented by IoU score.}
    \label{fig:think_vs_answer_supp}
\end{figure*}

\begin{figure*}[t]
    \centering
    \subfigure[]{
        \includegraphics[width=0.3\linewidth]{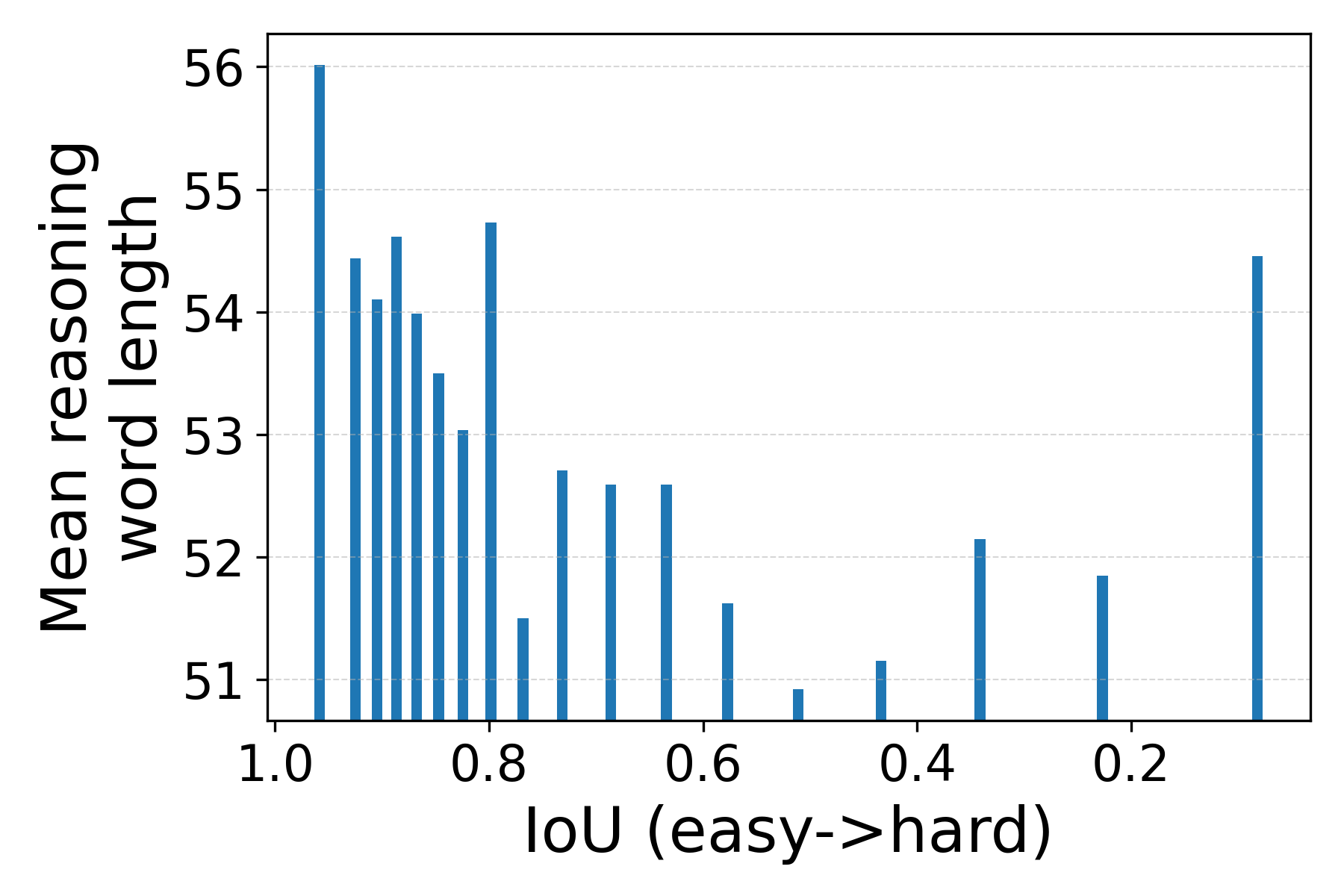}
        
    }
    \subfigure[]{
        \includegraphics[width=0.3\linewidth]{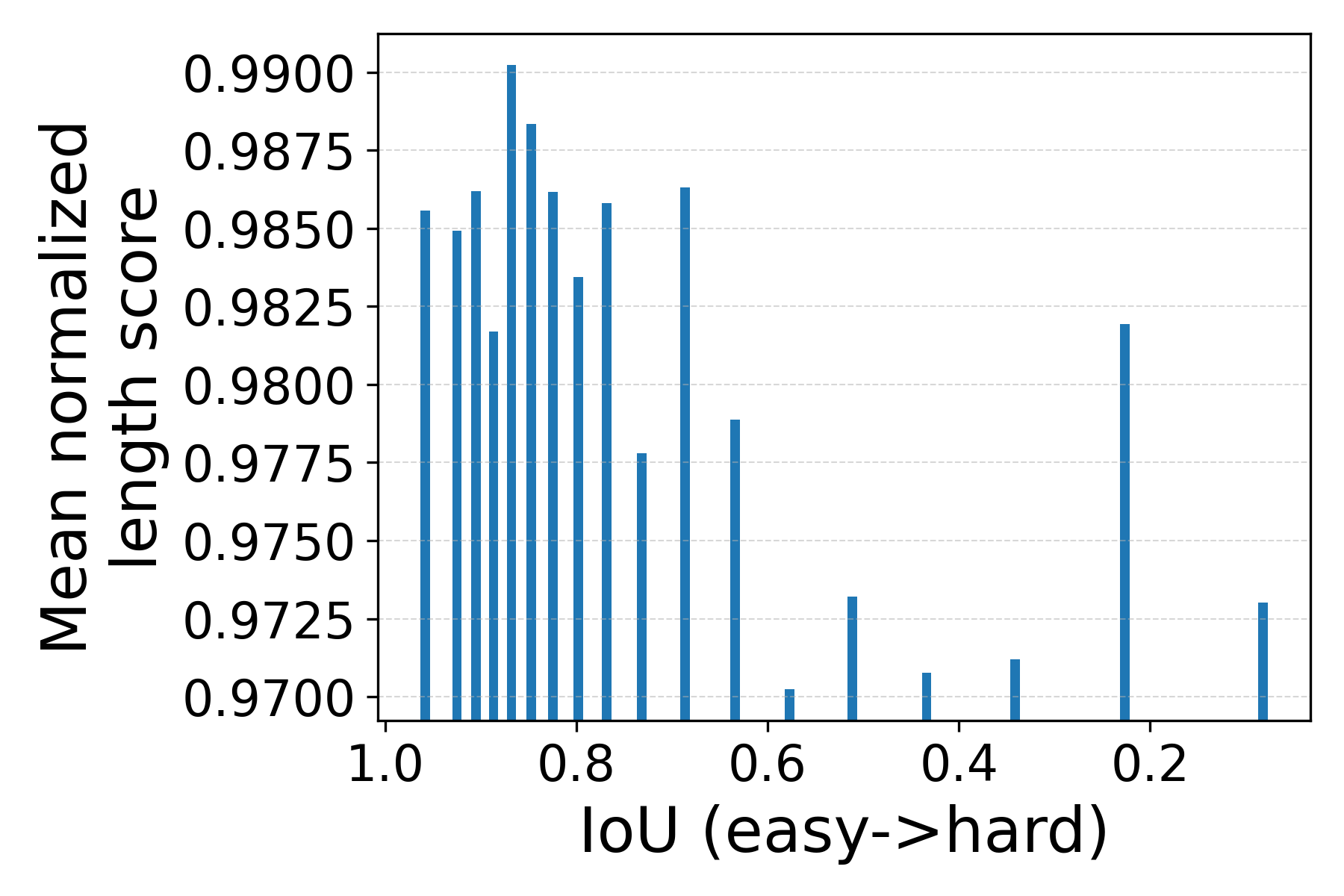}
        
    }
    \subfigure[]{
        \includegraphics[width=0.3\linewidth]{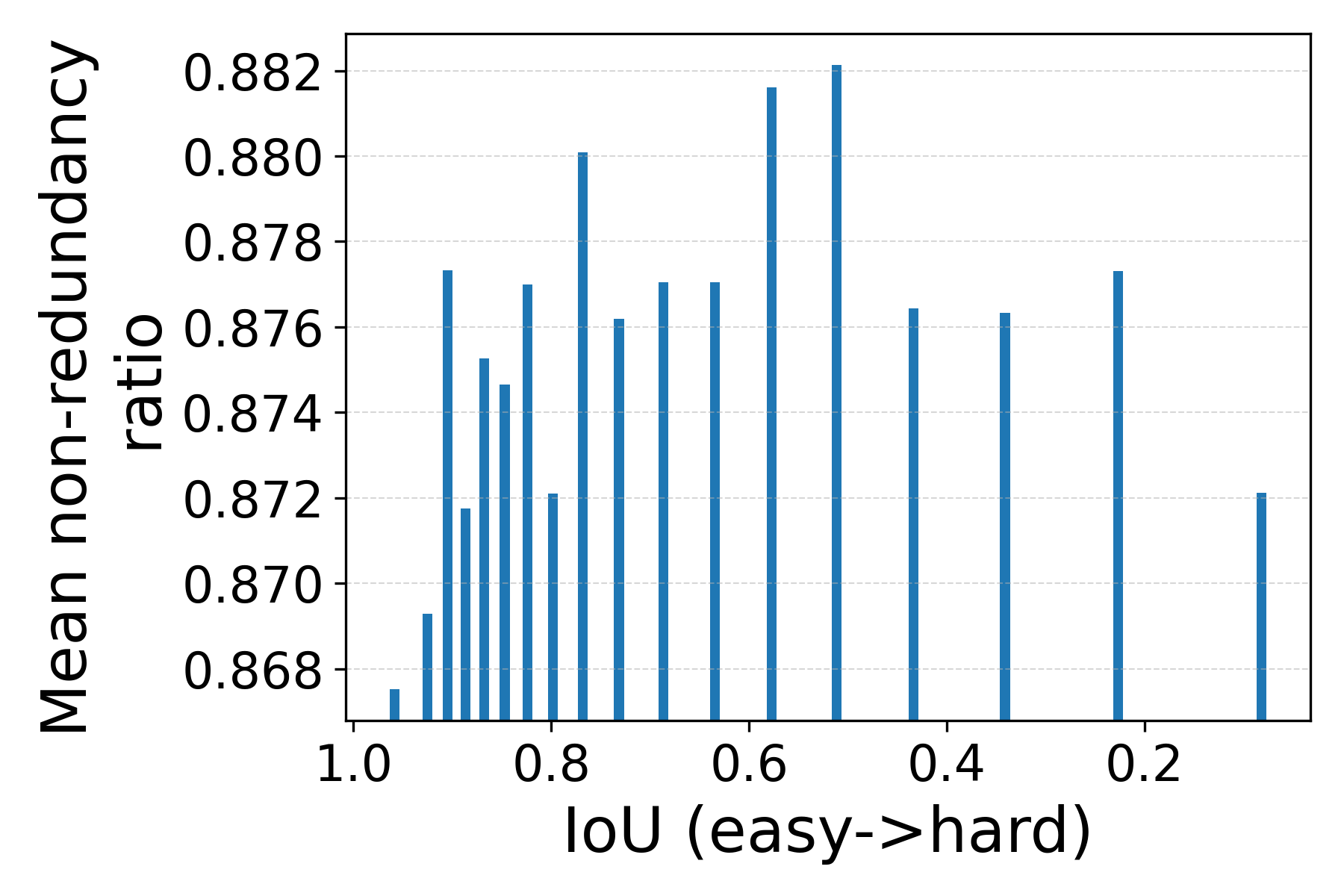}
        
    }
    \\
    \subfigure[]{
        \includegraphics[width=0.3\linewidth]{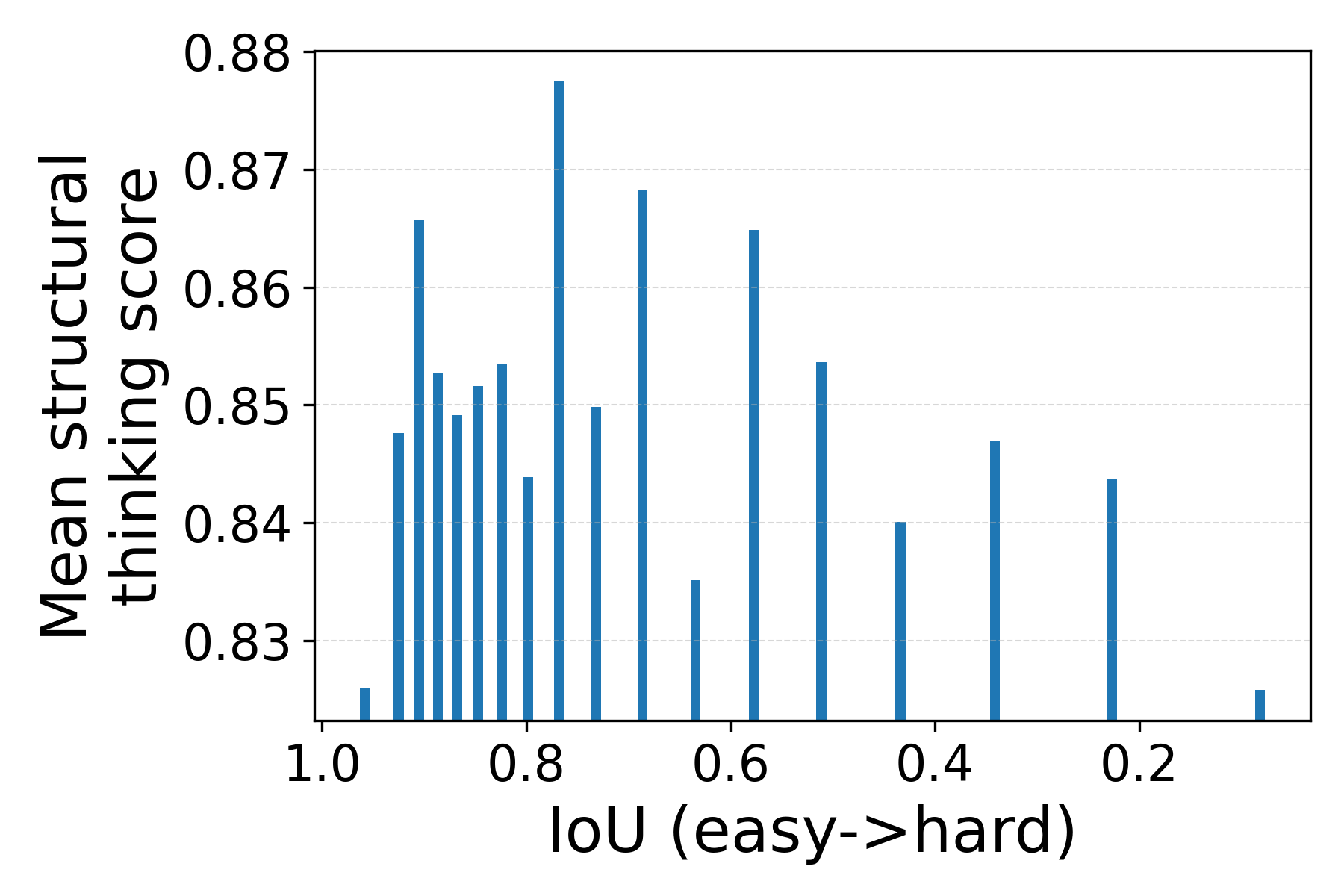}
        
    }
    \subfigure[]{
        \includegraphics[width=0.3\linewidth]{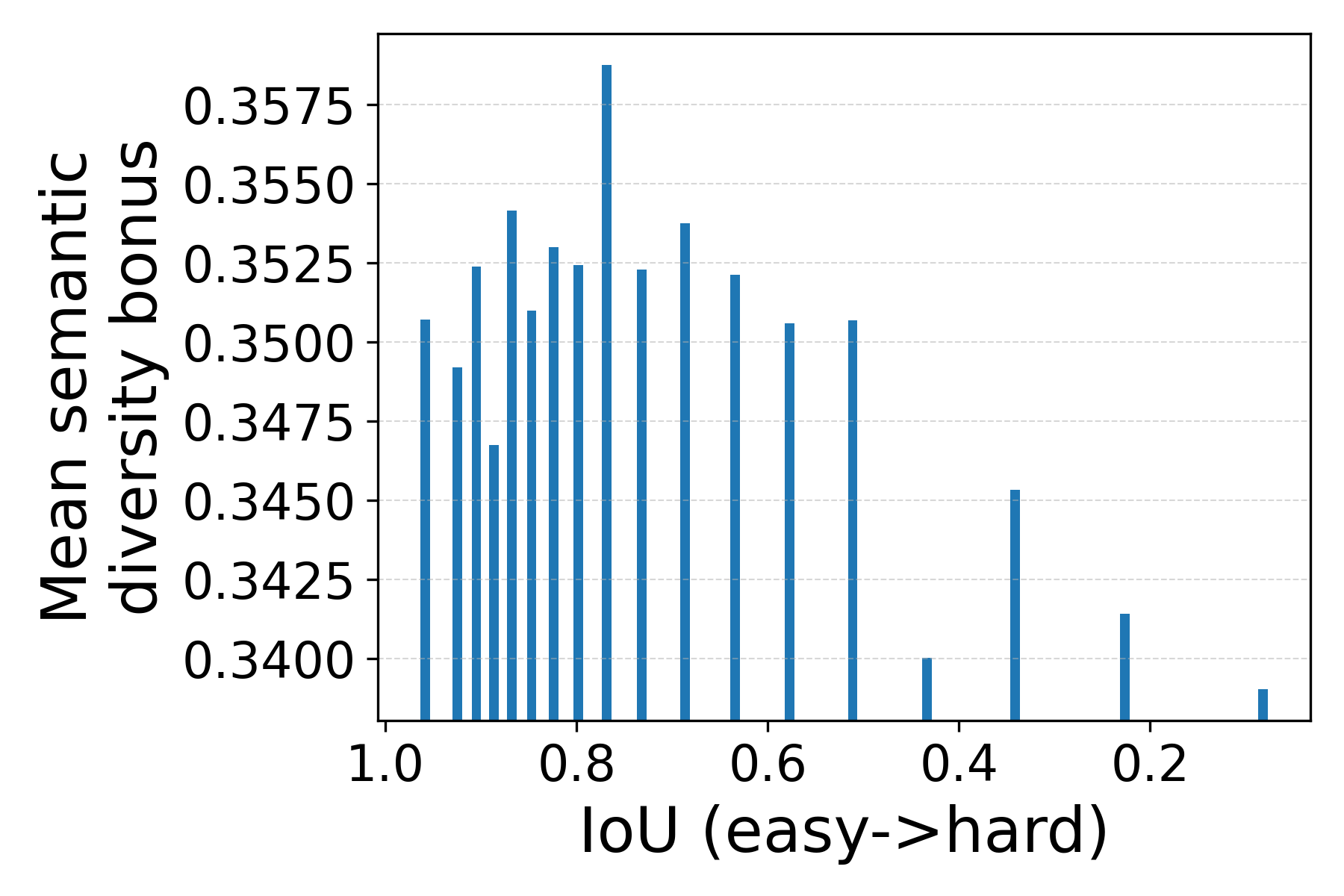}
        
    }
    \subfigure[]{
        \includegraphics[width=0.3\linewidth]{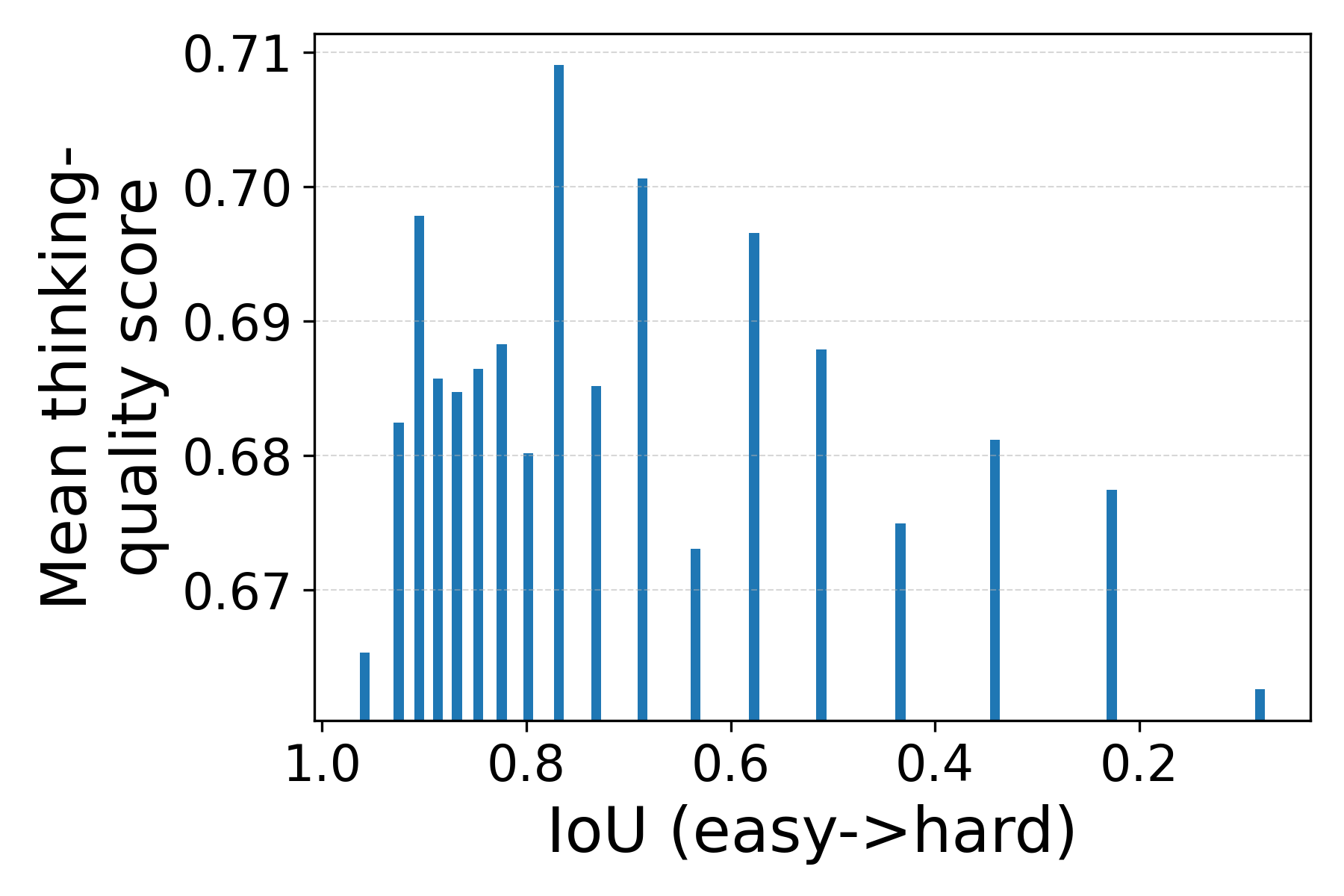}
        
    }
    \caption{The distribution of thinking-related variables with respect to accuracy, where the statistics are computed from GeoZero’s predictions on 7,500 samples from the DIOR-RSVG~\cite{dior-rsvg} test set. The sample accuracy is represented by the IoU score. The intervals are determined based on the quantiles of the IoU distribution, resulting in non-uniform spacing between the bars.}
    \label{fig:dist_reason_metric_acc}
\end{figure*}

    \item[\textbf{(3)}] \textbf{Structural Thinking.}  Figure~\ref{fig:think_vs_answer_supp}(d) clearly demonstrates that samples with low structural thinking scores~$q_t$ tend to have low average accuracy, whereas those with higher~$q_t$ consistently achieve better performance. This indicates that a well-structured reasoning process is a fundamental requirement for maintaining reliable predictions. However, the figure also shows that further increasing~$q_t$ does not lead to substantial performance gains, implying that, at this stage, the semantic content of reasoning (such as semantic diversity; see (e)) may play a more dominant role. Moreover, Figure~\ref{fig:think_vs_answer_supp}(j) shows that GeoZero produces well-structured reasoning chains for the majority of samples, reflecting its coherent reasoning capability.

    \item[\textbf{(4)}] \textbf{Semantic Diversity.} Figure~\ref{fig:think_vs_answer_supp}(e) further examines the semantic diversity bonus~$b_d$. We observe a non-monotonic relationship: moderate semantic diversity correlates with improved accuracy, as it reflects coherent multi-step reasoning. Once $b_d$ becomes excessively large, however, the reasoning often degenerates into fragmented or even contradictory statements, suggesting that the model is no longer performing genuine reasoning, which in turn hurts accuracy. Fortunately, Figure~\ref{fig:think_vs_answer_supp}(k) shows that for the vast majority of samples, the generated reasoning chains have $b_d$ values below~0.45, where semantic diversity plays a positive and constructive role.

    \item[\textbf{(5)}] \textbf{Overall Reasoning Quality.} Finally, Figures~\ref{fig:think_vs_answer_supp}(f) and (l) jointly indicate that GeoZero produces reasoning chains with high thinking-quality scores~($s_t$) for the majority of samples, which underpins the model’s overall performance across the test set. We also observe that some samples achieve good average accuracy despite relatively low~$s_t$ values. This phenomenon corresponds to the distinct leftmost bar in Figure~\ref{fig:think_vs_answer_supp}(j), where the structural thinking scores~($q_t$) are also low. This may be attributed to easier cases in which the model can make correct predictions without maintaining a highly organized reasoning structure, e.g., by exhibiting mild redundancy in expression, as seen in the third-to-last bar of Figure~\ref{fig:think_vs_answer_supp}(i).

\end{itemize}

In summary, these analyses reveal that \textbf{\textit{GeoZero consistently generates reasoning chains that align with the objectives of the training-time reward functions}}, including reasoning length, redundancy control, and structural organization, across the majority of samples in the DIOR-RSVG test set. This demonstrates that \textbf{\textit{the model achieves high-quality reasoning performance overall, thereby validating the effectiveness of the proposed approach.}}

\subsection{Reasoning Behavior across Samples.}

To further understand the reasoning behavior of GeoZero, we visualize the average values of reasoning metrics as a function of the sample prediction accuracy\footnote{Alternatively, one could employ frontier proprietary models to assess sample difficulty. However, due to the high API costs, we instead use GeoZero’s own prediction accuracy as a practical proxy.} (measured by IoU for grounding tasks), as shown in Figure~\ref{fig:dist_reason_metric_acc}. The bars are more densely distributed in the high-IoU regions, indicating that the test data is dominated by relatively easy samples. Nevertheless, a few highly challenging cases also exist, as evidenced by the rightmost bars corresponding to IoU values close to zero.

We next examine how GeoZero’s reasoning behavior varies with sample difficulty. For most samples (IoU $\in$ [0.8, 0.95]), the normalized length score $l_s(T)$ remains high, the number of reasoning words is relatively large, and the semantic diversity bonus~$b_d$ stays at an elevated level, while the non-redundancy ratio~$1-d(T)$, structural thinking score~$q_t$, and thinking-quality score~$s_t$ exhibit only minor fluctuations. These patterns suggest that the model can handle such cases in a stable manner: even though $1-d(T)$, $q_t$, and $s_t$ are not particularly high, GeoZero still engages in sufficiently thoughtful reasoning to produce accurate predictions.

Interestingly, for the easiest samples (IoU $\in$ [0.95, 1]), the model tends to produce even longer reasoning chains. This leads to mild redundancy, which slightly affects the reasoning-chain quality score $s_t$ through the structural thinking score~$q_t$, as reflected by the lower values of the leftmost bars in Figure~\ref{fig:dist_reason_metric_acc}(c), (d), and (f). However, since these samples are inherently simple, such redundancy does not compromise the model’s prediction accuracy.

When the sample difficulty slightly increases (IoU $\in$ [0.75, 0.8]), the model begins to “work harder.” This is reflected by an improvement in the structural thinking score~$q_t$ driven by a higher non-redundancy ratio~$1-d(T)$, as well as a further rise in the semantic diversity bonus~$b_d$. Together, these factors contribute to a higher thinking-quality score~$s_t$, indicating enhanced reasoning quality. The accompanying drop in accuracy is mainly attributable to task complexity rather than reduced reasoning ability, as the model actually demonstrates enhanced reasoning efforts.

As the sample difficulty further increases (IoU $\in$ [0.5, 0.75]), the reasoning capability of the model becomes challenged. We observe consistent declines in the reasoning length, normalized length score~$l_s(T)$, structural thinking score~$q_t$, semantic diversity~$b_d$, and thinking-quality score~$s_t$. This suggests that the model begins to lose its reasoning ability when confronted with difficult tasks, much like how humans may struggle to articulate coherent thoughts when facing particularly challenging problems.

When the sample difficulty increases even further (IoU~$<$~0.5), the reasoning-chain length starts to rise again, while the non-redundancy ratio~$1-d(T)$, structural thinking score~$q_t$, semantic diversity~$b_d$, and thinking-quality score~$s_t$ simultaneously decline. This pattern indicates a typical “reasoning collapse,” where the model produces repetitive and incoherent reasoning, effectively losing the ability to perform meaningful inference.

Nevertheless, it is worth emphasizing that even under the most challenging conditions, the average normalized length score $l_s(T)$ remains above~0.97, the average non-redundancy ratio $1-d(T)$ exceeds~0.86, the average structural thinking score $q_t$ is greater than~0.82, and the average semantic diversity $b_d$ stays within a narrow range of~0.33–0.35 with only minor variation, while the average thinking-quality score $s_t$ also remains above~0.66. These results demonstrate that \textbf{\textit{the reasoning quality across the vast majority of samples in all accuracy intervals still meets the expected standards}}, consistent with the findings in Figure~\ref{fig:think_vs_answer_supp}. Overall, these results suggest that \textbf{\textit{even as task difficulty increases, GeoZero maintains stable and meaningful reasoning performance.}}

Based on these observations, we believe that increasing the scale of challenging tasks during training could be a promising direction to further unlock the model’s reasoning potential. In particular, we observe that the model exhibits a distinct effort zone within the moderate difficulty range (IoU $\in$ [0.75, 0.8]), where it tends to work harder and achieve higher reasoning quality. These findings highlight the necessity of selecting training samples with appropriate difficulty, which could be key to consolidating and strengthening the model’s reasoning ability. Furthermore, given the mild redundancy observed in easy cases, it may be useful to introduce a reasoning efficiency regularization term into the reward design, encouraging the model to reason efficiently and avoid unnecessary cognitive effort on trivial tasks.

In summary, \textbf{\textit{GeoZero exhibits emergent reasoning behaviors that adapt to sample difficulty}}, leading to characteristic statistical patterns in these thinking-related variables. Analyzing such patterns helps us better understand the model’s reasoning dynamics and provides insights for designing more effective reasoning algorithms in the future.

\section{More Examples of Model Predictions}

\label{sec:visual_examples}

\begin{figure*}
    \centering
    \includegraphics[width=\linewidth]{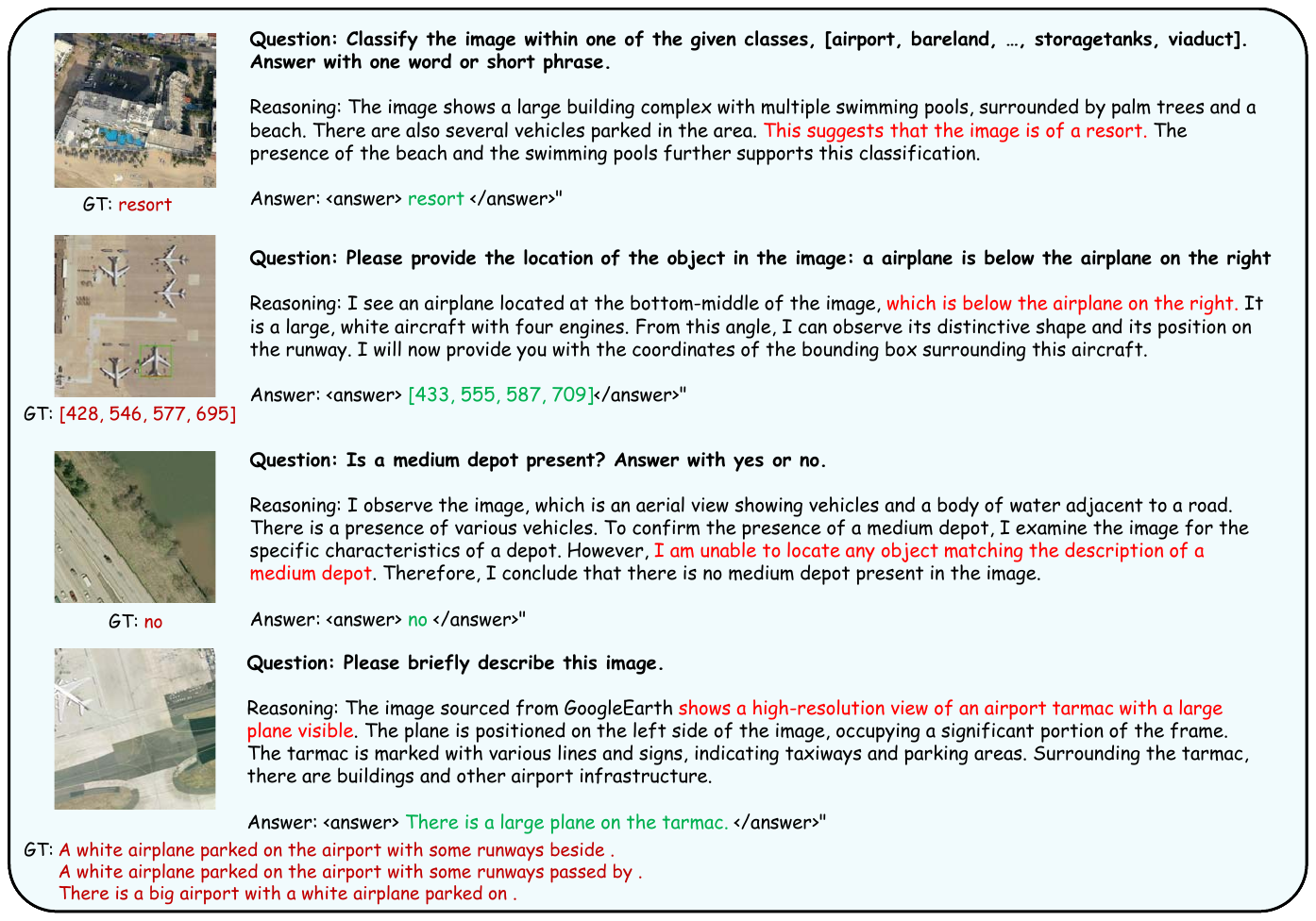}
    \caption{Examples of GeoZero outputs on different tasks.}
    \label{fig:vis_example}
\end{figure*}

We provide additional examples to visualize the model’s intermediate reasoning processes and final prediction results, as shown in Figure~\ref{fig:vis_example}.

\section{DataSheets}
\label{sec:datasheet}

\subsection{Motivation} 
\noindent \textbf{1. \textit{For what purpose was the dataset created?}} \vspace{0.5\baselineskip}\\
\noindent \textbf{A1:} GeoZero-Raw was constructed to facilitate the development of multimodal large language models that exhibit fully emergent reasoning on geospatial scenes. The dataset aims to help models acquire preliminary geospatial knowledge and stimulate universal reasoning capabilities across multiple remote sensing vision-language tasks.

\vspace{0.5\baselineskip}
\noindent \textbf{2. \textit{Who created the dataset (e.g., which team, research group) and on behalf of which entity (e.g., company, institution, organization)?}} \vspace{0.5\baselineskip}\\
\noindent\textbf{A2:} The dataset was created by the authors of the paper.

\vspace{0.5\baselineskip}
\noindent \textbf{3. \textit{Who funded the creation of the dataset?}} \vspace{0.5\baselineskip}\\
\noindent\textbf{A3:} The dataset creation was funded by the affiliations of the authors involved in this work.

\vspace{0.5\baselineskip}

\subsection{Composition}
\noindent \textbf{1. \textit{What do the instances that comprise the dataset represent (e.g., documents, photos, people, countries)? Are there multiple types of instances (e.g., movies, users, and ratings; people and interactions between them; nodes and edges)? Please provide a description.}} \vspace{0.5\baselineskip}\\
\noindent\textbf{A1:} The instances in the dataset consist of remote sensing images paired with their corresponding textual dialogues. These include multiple tasks such as scene classification, visual grounding, visual question answering, image captioning, object counting, multi-turn conversations, etc.

\vspace{0.5\baselineskip}
\noindent \textbf{2. \textit{How many instances are there in total (of each type, if appropriate)?}} \vspace{0.5\baselineskip}\\
\noindent\textbf{A2:} GeoZero-Raw contains 754,749 image-text pairs.

\vspace{0.5\baselineskip}
\noindent \textbf{3. \textit{Does the dataset contain all possible instances or is it a sample (not necessarily random) of instances from a larger set? If the dataset is a sample, then what is the larger set? Is the sample representative of the larger set (e.g., geographic coverage)? If so, please describe how this representativeness was validated/verified. If it is not representative of the larger set, please describe why not (e.g., to cover a more diverse range of instances, because instances were withheld or unavailable).}} \vspace{0.5\baselineskip}\\
\noindent\textbf{A3:} GeoZero-Raw is a curated subset of existing remote sensing vision and vision-language datasets (e.g., VHM-Instruct~\cite{pang2025vhm}, fMoW~\cite{fmow}). It samples diverse ground resolutions, sensor modalities, and scene categories to ensure representative geographic and semantic coverage, while the textual annotations are independently processed and refined by the authors.

\vspace{0.5\baselineskip}
\noindent \textbf{4. \textit{What data does each instance consist of? “Raw” data (e.g., unprocessed text or images) or features? In either case, please provide a description.}} \vspace{0.5\baselineskip}\\
\noindent\textbf{A4:} Each instance consists of two core components: remote sensing images and manually refined textual annotations. 

\vspace{0.5\baselineskip}
\noindent \textbf{5. \textit{Is there a label or target associated with each instance? If so, please provide a description.}} \vspace{0.5\baselineskip}\\
\noindent\textbf{A5:} Yes, each instance is associated with one or more textual question–answer pairs serving as supervision signals.

\vspace{0.5\baselineskip}
\noindent \textbf{6. \textit{Is any information missing from individual instances? If so, please provide a description, explaining why this information is missing.}} \vspace{0.5\baselineskip}\\
\noindent\textbf{A6:} No.

\vspace{0.5\baselineskip}
\noindent \textbf{7. \textit{Are relationships between individual instances made explicit (e.g., users’ movie ratings, social network links)? If so, please describe how these relationships are drawn.}} \vspace{0.5\baselineskip}\\
\noindent\textbf{A7:} Yes, explicit relationships between individual instances are defined based on task descriptors at the beginning of the instruction text, which indicate whether instances belong to the same task type.

\vspace{0.5\baselineskip}
\noindent \textbf{8. \textit{Are there recommended data splits (e.g., training, development/validation, testing)? If so, please provide a description of these splits, explaining the rationale behind them.}} \vspace{0.5\baselineskip}\\
\noindent\textbf{A8:} The dataset is divided into two subsets: GeoZero-Instruct and GeoZero-Hard, through hard-sample selection and image deduplication. These subsets are used for supervised fine-tuning and reinforcement learning, respectively, to activate geospatial reasoning from scratch.

\vspace{0.5\baselineskip}
\noindent \textbf{9. \textit{Are there any errors, sources of noise, or redundancies in the dataset? If so, please provide a description.}} \vspace{0.5\baselineskip}\\
\noindent\textbf{A9:} No.

\vspace{0.5\baselineskip}
\noindent \textbf{10. \textit{Is the dataset self-contained, or does it link to or otherwise rely on external resources?}} \vspace{0.5\baselineskip}\\
\noindent\textbf{A10:} The dataset is self-contained in terms of structure and usability. All necessary textual annotations required for training are included. The image data originates from publicly available remote sensing datasets (e.g., VHM-Instruct~\cite{pang2025vhm}, fMoW~\cite{fmow}). Users can access the images via the provided links.

\vspace{0.5\baselineskip}
\noindent \textbf{11. \textit{Does the dataset contain data that might be considered confidential?}} \vspace{0.5\baselineskip}\\
\noindent\textbf{A11:} No.

\vspace{0.5\baselineskip}
\noindent \textbf{12. \textit{Does the dataset contain data that, if viewed directly, might be offensive or distressing?}} \vspace{0.5\baselineskip}\\
\noindent\textbf{A12:} No.

\vspace{0.5\baselineskip}

\subsection{Collection Process} 
\noindent \textbf{1. \textit{How was the data associated with each instance acquired?}} \vspace{0.5\baselineskip}\\
\noindent\textbf{A1:} The images in each instance were obtained from existing publicly available datasets, while the textual annotations were generated by us through instruction formatting.

\vspace{0.5\baselineskip}
\noindent \textbf{2. \textit{What mechanisms or procedures were used to collect the data?}} \vspace{0.5\baselineskip}\\
\noindent\textbf{A2:} The images in the dataset were directly downloaded from existing publicly available datasets, while the textual annotations were generated using Python scripts written by us for format conversion.

\vspace{0.5\baselineskip}
\noindent \textbf{3. \textit{If the dataset is a sample from a larger set, what was the sampling strategy?}} \vspace{0.5\baselineskip}\\
\noindent\textbf{A3:} The dataset is initially built upon the VHM-Instruct dataset~\cite{pang2025vhm}. Since we mainly focus on scene classification, visual grounding, visual question answering, and image captioning tasks, we further supplement additional samples from the training sets of existing datasets for each task. This process alleviates the task imbalance issue in VHM-Instruct and ensures comprehensive task coverage across different modalities.

\vspace{0.5\baselineskip}
\noindent \textbf{4. \textit{Who was involved in the data collection process, and how were they compensated?}} \vspace{0.5\baselineskip}\\
\noindent\textbf{A4:} The data collection and verification were conducted by the authors. No external parties or contractors were involved, and the authors were not compensated beyond their academic roles and institutional affiliations.

\vspace{0.5\baselineskip}
\noindent \textbf{5. \textit{Over what timeframe was the data collected?}} \vspace{0.5\baselineskip}\\
\noindent\textbf{A5:} The data collection process, including download, selection, and instruction formatting, took approximately 2 weeks in total.

\vspace{0.5\baselineskip}

\subsection{Preprocessing/Cleaning/Labeling} 
\noindent \textbf{1. \textit{Was any preprocessing, cleaning, or labeling performed?}} \vspace{0.5\baselineskip}\\
\noindent\textbf{A1:} Each instance undergoes preprocessing that includes instruction formatting, such as specifying task descriptors, randomly selecting textual hints, and performing coordinate system transformations for grounding.

\vspace{0.5\baselineskip}
\noindent \textbf{2. \textit{Was the raw data saved in addition to the processed data?}} \vspace{0.5\baselineskip}\\
\noindent\textbf{A2:} No.

\vspace{0.5\baselineskip}
\noindent \textbf{3. \textit{Is the preprocessing software available?}} \vspace{0.5\baselineskip}\\
\noindent\textbf{A3:} The preprocessing was implemented in Python.

\vspace{0.5\baselineskip}

\subsection{Uses} 
\noindent \textbf{1. \textit{Has the dataset been used for any tasks already?}} \vspace{0.5\baselineskip}\\
\noindent\textbf{A1:} No.

\vspace{0.5\baselineskip}
\noindent \textbf{2. \textit{Is there a repository linking related works using this dataset?}} \vspace{0.5\baselineskip}\\
\noindent\textbf{A2:} N/A.

\vspace{0.5\baselineskip}
\noindent \textbf{3. \textit{What other tasks could the dataset be used for?}} \vspace{0.5\baselineskip}\\
\noindent\textbf{A3:} GeoZero-Raw can be used for remote sensing vision-language tasks, such as scene classification, visual grounding, visual question answering, and image captioning.

\vspace{0.5\baselineskip}
\noindent \textbf{4. \textit{Is there anything about the dataset’s composition that might impact future uses?}} \vspace{0.5\baselineskip}\\
\noindent\textbf{A4:} No.

\vspace{0.5\baselineskip}
\noindent \textbf{5. \textit{Are there tasks for which the dataset should not be used?}} \vspace{0.5\baselineskip}\\
\noindent\textbf{A5:} No.

\vspace{0.5\baselineskip}

\subsection{Distribution} 
\noindent \textbf{1. \textit{Will the dataset be distributed to third parties?}} \vspace{0.5\baselineskip}\\
\noindent\textbf{A1:} Yes, it will be publicly available.

\vspace{0.5\baselineskip}
\noindent \textbf{2. \textit{How will the dataset be distributed?}} \vspace{0.5\baselineskip}\\
\noindent\textbf{A2:} It will be released via an online repository, accessible through GitHub and HuggingFace.

\vspace{0.5\baselineskip}
\noindent \textbf{3. \textit{When will the dataset be distributed?}} \vspace{0.5\baselineskip}\\
\noindent\textbf{A3:} The dataset will be publicly available upon acceptance of the paper.

\vspace{0.5\baselineskip}
\noindent \textbf{4. \textit{Under what license will it be distributed?}} \vspace{0.5\baselineskip}\\
\noindent\textbf{A4:} Creative Commons Attribution-NonCommercial-ShareAlike 4.0 License.

\vspace{0.5\baselineskip}
\noindent \textbf{5. \textit{Have any third parties imposed IP-based restrictions?}} \vspace{0.5\baselineskip}\\
\noindent\textbf{A5:} No.

\vspace{0.5\baselineskip}
\noindent \textbf{6. \textit{Do any export controls or other regulatory restrictions apply?}} \vspace{0.5\baselineskip}\\
\noindent\textbf{A6:} No.

\vspace{0.5\baselineskip}

\subsection{Maintenance} 
\noindent \textbf{1. \textit{Who will maintain the dataset?}} \vspace{0.5\baselineskip}\\
\noindent\textbf{A1:} The authors.

\vspace{0.5\baselineskip}
\noindent \textbf{2. \textit{How can the maintainers be contacted?}} \vspace{0.5\baselineskip}\\
\noindent\textbf{A2:} E-mail.

\vspace{0.5\baselineskip}
\noindent \textbf{3. \textit{Is there an erratum?}} \vspace{0.5\baselineskip}\\
\noindent\textbf{A3:} No known errors or issues have been identified to date.

\vspace{0.5\baselineskip}
\noindent \textbf{4. \textit{Will the dataset be updated?}} \vspace{0.5\baselineskip}\\
\noindent\textbf{A4:} Currently, there are no plans for regular updates. However, if significant improvements or extensions are made, future revisions of the dataset may be released.

\vspace{0.5\baselineskip}
\noindent \textbf{5. \textit{Will older versions be maintained?}} \vspace{0.5\baselineskip}\\
\noindent\textbf{A5:} Previous versions will be preserved for reproducibility.

\vspace{0.5\baselineskip}
\noindent \textbf{6. \textit{If others want to extend or contribute, is there a mechanism?}} \vspace{0.5\baselineskip}\\
\noindent\textbf{A6:} Yes, we will provide dataset processing scripts to facilitate future extensions and contributions.


\end{document}